\pdfoutput=1
\documentclass[11pt]{article}

\usepackage[final]{acl}

\usepackage{times}
\usepackage{latexsym}

\usepackage[T1]{fontenc}
\usepackage[utf8]{inputenc}

\usepackage{microtype}

\usepackage{inconsolata}

\usepackage{graphicx,adjustbox}
\usepackage{caption}
\usepackage{threeparttable}
\usepackage{subcaption}
\usepackage{tabularx}
\usepackage{float}

\usepackage{xcolor}         
\usepackage{colortbl}         
\usepackage{amssymb} 
\usepackage{enumitem}
\newenvironment{itemize*}%
  {\begin{itemize}%
    \setlength{\itemsep}{0.9pt}%
    \setlength{\parskip}{0.9pt}%
    \setlength{\topsep}{0.9pt}}%
  {\end{itemize}}

\newenvironment{enumerate*}%
  {\begin{enumerate}%
    \setlength{\itemsep}{0.9pt}%
    \setlength{\parskip}{0.9pt}%
    \setlength{\topsep}{0.9pt}}%
  {\end{enumerate}}

\newenvironment{rqs}%
  {\begin{enumerate}[label={\textbf{RQ\arabic*}},wide=0pt]%
    \setlength{\itemsep}{0.9pt}%
    \setlength{\parskip}{0.9pt}%
    \setlength{\topsep}{0.9pt}}%
  {\end{enumerate}}

\usepackage{todonotes}
\makeatletter
\newcommand*\iftodonotes{\if@todonotes@disabled\expandafter\@secondoftwo\else\expandafter\@firstoftwo\fi} 
\makeatother

\usepackage[nameinlink]{cleveref}

\crefformat{section}{\S#2#1#3} 
\crefformat{subsection}{\S#2#1#3}
\crefformat{subsubsection}{\S#2#1#3}

\title{Research Community Perspectives \\ on ``Intelligence'' and Large Language Models}

\author{
 \textbf{Bertram Højer\textsuperscript{1}},
 \textbf{Terne Sasha Thorn Jakobsen\textsuperscript{2}},
 \textbf{Anna Rogers\textsuperscript{1}},
 \textbf{Stefan Heinrich\textsuperscript{1}}
\\
 \textsuperscript{1}IT University of Copenhagen,
 \textsuperscript{2}University of Copenhagen
\\
 \small{
    \textbf{GitHub:} \href{https://github.com/bertramhojer/perspectives-on-intelligence}{bertramhojer/perspectives-on-intelligence}
 }
}

\begin{document}
\maketitle

\begin{abstract}
Despite the widespread use of ``artificial intelligence'' (AI) framing in Natural Language Processing (NLP) research, it is not clear what researchers mean by ``intelligence''. To that end, we present the results of a survey on the notion of ``intelligence'' among researchers and its role in the research agenda. The survey elicited complete responses from 303 researchers from a variety of fields including NLP, Machine Learning (ML), Cognitive Science, Linguistics, and Neuroscience.
We identify $3$ criteria of intelligence that the community agrees on the most: \textit{generalization, adaptability, \& reasoning}.
Our results suggests that the perception of the current NLP systems as ``intelligent'' is a minority position (29\%).
Furthermore, only 16.2\% of the respondents see developing intelligent systems as a research goal, and these respondents are more likely to consider the current systems intelligent.
\end{abstract}
\section{Introduction}
\label{sec:intro}

Whether machines can be considered intelligent has long been a question of interest in both academia and industry. Researchers have a rich tradition of devising intelligence tests for machines, starting from the Turing test \cite{turingICOMPUTINGMACHINERYINTELLIGENCE1950}, and the industry increasingly markets ``smart'' and ``intelligent'' systems. A recent iteration hereof are LLM-based systems, also referred to as ``generative AI'',\footnote{We are only concerned with systems based on Large Language Models (LLMs), i.e., models trained on large text corpora that are used for transfer learning \cite{RogersLuccioni_2024_Position_Key_Claims_in_LLM_Research_Have_Long_Tail_of_Footnotes}. We use the term ``AI'' in line with its common use in the field, but we note that it entails that the term ``intelligence'' is applicable, which, as our results suggest, is not accepted by the majority of the respondents.} which have been suggested to show ``sparks of artificial general intelligence'' \cite{bubeckSparksArtificialGeneralIntelligenceEarly2023}.

\begin{figure}[!t]
    \centering
    \includegraphics[width=.95\linewidth]{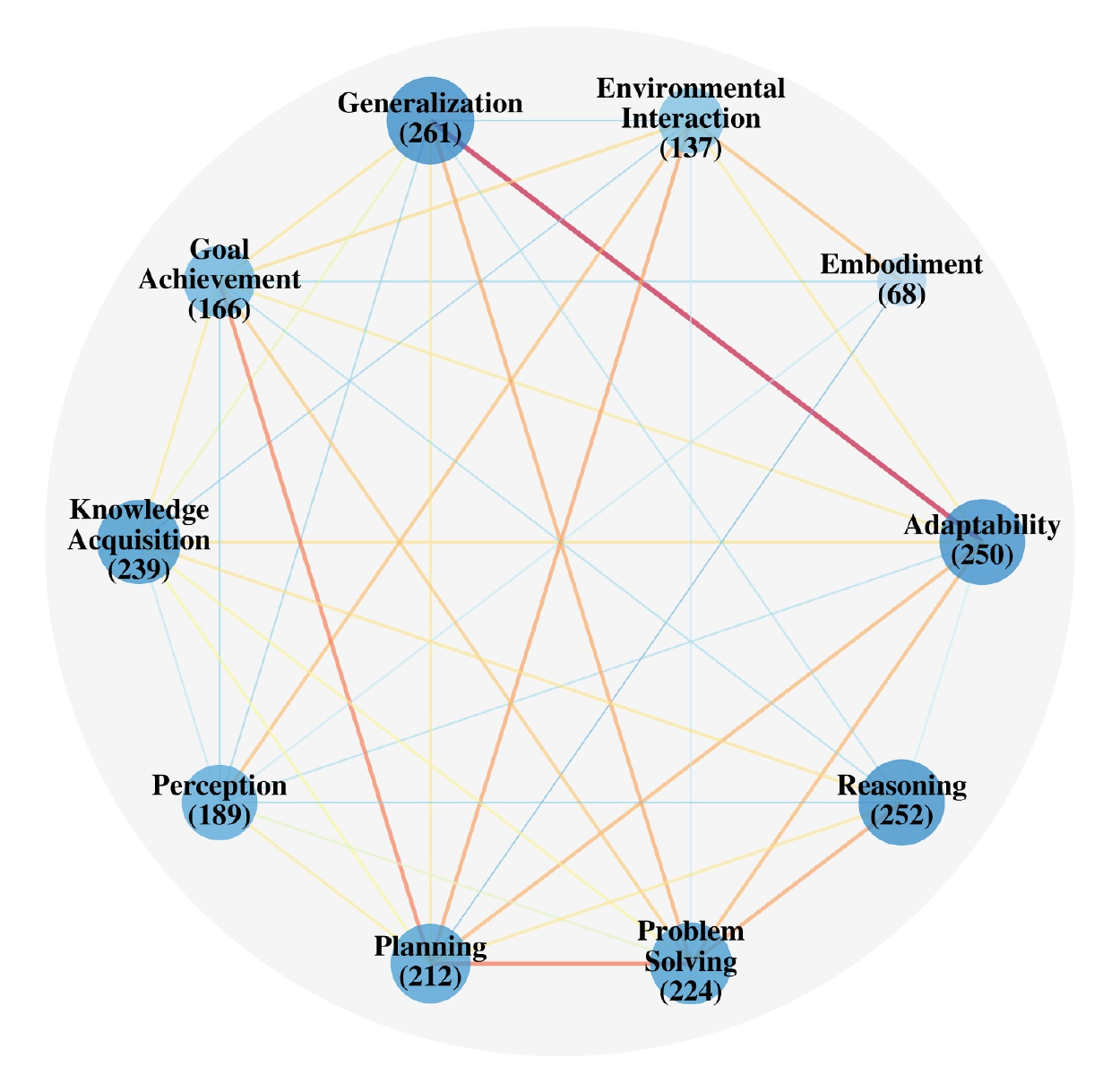}
    \caption{Correlation between criteria that the survey respondents selected as relevant for their notion of ``intelligence''. Darker edges indicate stronger correlations, larger nodes indicate higher relevance. Only edges with $\phi > |0.1|$ are shown.}
    \label{fig:criteria-network-graph}
\end{figure}

Irrespective of one's position on whether systems based on current ML technology can be considered intelligent, there is undeniably an increase in the phenomenon we refer to as \textit{seeming intelligence}: the user's perception that a system is intelligent, based on their interaction with that system. This perception arises from the Eliza effect: humans have a strong tendency to attribute to their conversational partners, even simple chatbots, ``all sorts of background knowledge, insights and reasoning ability'' \cite{weizenbaumELIZAComputerProgramStudyNatural1966}, and the perception of intelligence is a driver of anthropomorphization, irrespective of whether a system is represented via an avatar or not \cite{kimAnthropomorphicResponseUnderstandingInteractionsHumans2023}.

The tendency to anthropomorphize AI systems is not limited to the lay public or marketing by tech companies:\footnote{OpenAI describe their base class of models as "\textit{(...) models without instruction following that can \textbf{understand} (...) natural language and code}", GPT-4 is better "\textit{... thanks to its broader general \textbf{knowledge} and advanced \textbf{reasoning} capabilities."} and the o1 class of models "\textit{\textbf{think} before they answer}".}
%\cite{OpenAIPlatform}. Anthropic stated that the fundamental limitation of "\textbf{logical reasoning}" has already been solved with current systems \cite{anthropicaiCoreViewsAISafetyWhen2023}.}
\citet{mitchellWhyAIHarderWeThink2021} points out that even researchers mislead themselves by using terms like ``understanding'', which have long been criticized as ``wishful mnemonics'' \cite{mcdermottArtificialIntelligenceMeetsNaturalStupidity1976}. In particular, we have a longstanding tradition of evaluating ``understanding'' \cite[e.g.][]{wangGLUEMultiTaskBenchmarkAnalysisPlatform2018, wangSuperGLUEStickierBenchmarkGeneralPurposeLanguage2020}, ``reasoning'' \cite[e.g.][]{vilaresHEADQAHealthcareDatasetComplexReasoning2019, srivastava2023beyond}, ``behaviors'' \cite[e.g.][]{changLanguageModelBehaviorComprehensiveSurvey2023} and ``capabilities'' \cite[e.g.][]{chen2024llmarenaassessingcapabilitieslarge,srivastava2023beyond}.

\medskip 

Given the above tendency towards anthropomorphization and the long history of ``wishful mnemonics'' in the academic literature, the goal of this study is to provide a snapshot of how researchers think about ``intelligence'' two years after the release of ChatGPT \cite{OpenAI_2022_Introducing_ChatGPT}.
We pose the following research questions:
\begin{rqs}
    \item \label{rq:def} \textit{What are key criteria for a system to be considered ``intelligent'', and to what degree do researchers in different fields, occupations, and career stages agree?} 
    \item \label{rq:chatgpt} \textit{What is the perception of the ``intelligence'' of current LLM-based systems such as ChatGPT?}
    \item \label{rq:agenda} \textit{What role, if any, does the research agenda of respondents play in the notion of ``intelligence''?}
\end{rqs}

Our results indicate a high level of coherence across groups in terms of the criteria that are relevant for intelligence. Independent of one's field of research, occupation, and career stage, most researchers believe that \textit{generalization}, \textit{adaptability}, and \textit{reasoning} are key aspects of intelligence (\ref{rq:def}, illustrated in \autoref{fig:criteria-network-graph}). We additionally observe a general skepticism regarding the attribution of intelligence to LLM-based systems, although it decreases for ``future systems based on a similar technology'' (\ref{rq:chatgpt}). For \ref{rq:agenda}, we find that only a small subset of respondents consider it their goal to build intelligent technology, with the majority rather aiming to add to the scientific knowledge. Interestingly, the survey shows that those who consider `creating intelligent systems' as their research goal are also more likely to attribute intelligence to the current systems (\ref{rq:chatgpt}, \ref{rq:agenda}).
 
\section{Background and Related Work}
\label{sec:related}

\subsection{Survey methodology}

%To study attitudes and beliefs is to attempt to observe the unobservable. There are situations where attitudes become more evident, e.g. voting in elections, but in the absence of such events, 
When beliefs are implicit rather than clearly expressed in words or actions, administering surveys is often the only way of studying the attitudes of a large set of people \cite{robsonRealWorldResearchResourceSocial1999}. 
This methodology is by no means perfect: the challenges of online self-administered surveys include low response rates and the possibility of misunderstandings, inaccurate self-reporting, and responses not made in good faith. %Furthermore, the reporting of one's beliefs is challenging: it is affected by the respondent's current circumstances and might not always be accurate. 
%
%Yet, to answer our research questions, the online survey is the most appropriate method. In the context of AI research, where definitions are often contested and subjective, 
%
Still, online surveys are a widely used methodology that allows us to systematically capture the beliefs held by a broad set of researchers, in different fields, occupations, and career stages, and analyze responses statistically \cite{phillipsProperApplicationsSurveysStudyMethodology2017}. 

Among the prior surveys in the NLP community are the meta-survey of researcher beliefs \cite{michaelWhatNLP2023} and the survey of views on the factors important for paper-reviewer matching \cite{jakobsenWhatFactorsShouldPaperReviewer2022}. These surveys were distributed similarly to ours and received 327 and 180 valid responses respectively.

Surveys have also been deployed more broadly to study researchers' beliefs about the current state of AI, but these have mostly focused on impact predictions rather than beliefs about the actual technology. For example, the recent Pew survey on how US adults and experts view AI \cite{pasquiniHowUSPublicAIExperts2025} does not elicit how respondents define or use that term. Neither are definitions the focus of `timeline to AGI' surveys.\footnote{\url{https://wiki.aiimpacts.org/ai_timelines/predictions_of_human-level_ai_timelines/ai_timeline_surveys/ai_timeline_surveys}} Some of the aspects that we consider in this work were also in the AAAI 2025 presidential panel \cite{rossiFutureAIResearch2025}.

\subsection{Defining ``Intelligence''}

\begin{table*}[!t]
\centering
\resizebox{\textwidth}{!}{\begin{tabular}{l l r}
\hline
\textbf{Criterion} & \textbf{Survey definition} & \textbf{Sources} \\
\hline
Prob. solving & \textit{solving specific problems in a familiar domain} & \citet{turingICOMPUTINGMACHINERYINTELLIGENCE1950, mitchellWhyAIHarderWeThink2021}  \\
Knowledge acqusition & \textit{learning, understanding or gaining knowledge and skills} & \citet{Piaget1952OriginsIntelligence,Chollet2019MeasureIntelligence} \\
Adaptability & \textit{making sense of new environments and/or handling novel tasks} & \citet{leggCollectionDefinitionsIntelligence2007, Pfeifer2006BodyIntelligence} \\
Generalization & \textit{successfully handling new types of data and situations} & \citet{Marcus2003AlgebraicMind, Tenenbaum2011Mind}  \\
Goal achievement &\textit{accomplishing defined objectives and optimizing for specific tasks} & \citet{Piaget1952OriginsIntelligence, VanGerven2017ComputationalIntelligence} \\
Reasoning & \textit{logical inference - deductive, abductive etc.} & \citet{leggCollectionDefinitionsIntelligence2007, LakeTenenbaum2017LearnThinkPeople, dennettBacteriaBachBackEvolutionMinds2018} \\
% Understanding & \citet{Gardner1983Mind, Tenenbaum2011Mind} \\
Perception & \textit{extracting and acting upon useful information from the environment} & \citet{Smith2005development, VanGerven2017ComputationalIntelligence} \\
Planning & \textit{anticipating future events and organizing actions based on a deliberate strategy} & \citet{albusOutlineTheoryIntelligence1991} \\
% Creativity & \citet{Wiggins2020CreativityInformationConsciousness} \\
Environment interaction &\textit{embedded in physical or virtual environment} & \citet{Piaget1952OriginsIntelligence, Pfeifer2001UnderstandingIntelligence} \\
% Consciousness & \citet{Wiggins2020CreativityInformationConsciousness} \\
Embodiment & \textit{being situated in a physical environment} & \citet{Smith2005development, Pfeifer2006BodyIntelligence} \\
\hline
\end{tabular}}
\caption{The criteria of ``intelligence'' presented to the survey participants, and their sources. Each criteria was provided alongside a simple working definition.}
\label{tab:criteria-definition}
\end{table*}

``Intelligence'' is studied in multiple fields, and a complete account of all its aspects is beyond the scope of this work.
Our starting point was a list of $72$ definitions from psychology, dictionaries, and AI research, collected by \citet{leggCollectionDefinitionsIntelligence2007}. While not representative of all research on intelligence  (esp. Philosophy and Neuroscience), this list has been influential in the AI/ML community, and is thus a good source of definitions that are known in both psychology and AI/ML. They emphasize two key aspects of intelligence: it is (a) ascribed to agents in the context of and environment, and (b) concerns the ability of a system to adapt to novel situations. 

While these aspects are a common denominator of the definitions, they are not exhaustive. We conducted additional analysis of the described list, to identify other concepts that are commonly mentioned in the current ML/AI literature. We also consulted other relevant literature (in particular, \citet{turingICOMPUTINGMACHINERYINTELLIGENCE1950, albusOutlineTheoryIntelligence1991, leggFormalMeasureMachineIntelligence2006, Tenenbaum2011Mind, LakeTenenbaum2017LearnThinkPeople, VanGerven2017ComputationalIntelligence, dennettBacteriaBachBackEvolutionMinds2018} - see \autoref{tab:criteria-definition}).

Most concepts relevant to intelligence are themselves fuzzy and could be understood differently by people in different fields. We accompanied each criteria by a brief working definition, more succinct than a typical research definition. These are shown in \autoref{tab:criteria-definition}.\footnote{The survey also included 3 criteria (language understanding, creativity, consciousness) for which the participants were invited to use their own notion of the term. The analysis including these criteria is available in \autoref{app:all-criteria}.} We make no claim that this set of criteria is exhaustive or necessarily representative of all intelligence research, and we invite future work to experiment with different sets of criteria. However, no study attempting to elucidate the notion of ``intelligence'' can likely ever claim to be exhaustive. This is due to the inherent complexity of the concept and the difficulty we observe for researchers in considering even a small set of criteria.

\subsection{Relevant NLP Research Directions}

There are many research directions relevant to notions of ``intelligence'' which we cannot do justice in scope of this work. The remainder of this section provides pointers to three prominent directions within the field of NLP. 

\paragraph{The meaning debate.} There is an active debate on whether terms like ``meaning'' and ``understanding'' apply to LLMs, with prominent opponents \cite{bender-koller-2020-climbing,merrill-etal-2021-provable} and proponents \cite{li-etal-2024-vision-language}.

\paragraph{``Science of datasets''.} Benchmarks such as SuperGLUE \cite{wangSuperGLUEStickierBenchmarkGeneralPurposeLanguage2020}, MMLU \cite{hendrycksMeasuringMassiveMultitaskLanguageUnderstanding2021}, and GSM8K \cite{cobbeTrainingVerifiersSolveMathWord2021} are used both for academic evaluation of LLM-based systems, and in their marketing.
%But how much we can deduce from the current benchmark results is up for debate.
Ideally, the creation of benchmarks and training resources should consider the construct validity of the phenomena being modeled, and hence rely on specific definitions of their target phenomena and representativeness of their samples \cite{schlangen-2021-targeting,rajiAIEverythingWholeWideWorld2021}, but we do not yet have solid methodology for doing this well for complex and underdefined constructs like ``reasoning''. Related research directions include documenting \cite{BenderFriedman_2018_Data_Statements_for_Natural_Language_Processing_Toward_Mitigating_System_Bias_and_Enabling_Better_Science,GebruMorgensternEtAl_2020_Datasheets_for_Datasets} and `measuring' the existing datasets \cite{dodge-etal-2021-documenting,mitchell2023measuringdata}, in particular to identify shortcuts in them \cite{gururangan-etal-2018-annotation,gardner-etal-2021-competency} and measure test contamination \cite{deng-etal-2024-investigating,dong-etal-2024-generalization,jacovi-etal-2023-stop}. Another relevant direction is creating adversarial datasets to `catch' the models relying on shallow heuristics \cite[e.g.][]{mccoy-etal-2019-right}. 

\paragraph{Interpretability research.} The field of interpretability \cite{belinkov-glass-2019-analysis,rogers-etal-2020-primer}, including the mechanistic camp \cite{saphra-wiegreffe-2024-mechanistic,olssonIncontextLearningInductionHeads2022,nandaProgressMeasuresGrokkingMechanisticInterpretability2023}, aims to identify the internal mechanisms of the models through which they arrive at their predictions. Assuming that we know what the ``correct'' reasoning steps should be, this could allow us to check whether the model adheres to them, or relies on shallow heuristics \cite[e.g.][]{raychoudhuryMachineReadingFastSlowWhen2022}.

\section{Survey Structure and Distribution}
\label{sec:methodology}

The survey was developed using the \href{https://www.survey-xact.dk/}{SurveyXact} platform. We initially provide the introductory information regarding anonymity and informed consent, followed by questions related to basic demographic information relevant for the analysis. This is followed by questions pertaining to the respondents' perspectives on the notion of ``intelligence''. Responses were collected between November 2024 and January 2025. The survey form and response data (with coarse-grained geographic data for increased anonymity) is provided on the accompanying github repository, and a full list of questions can be found in \autoref{app:sec:questions}.

%\subsection{Survey Distribution}
The survey was distributed through social media (Bluesky and LinkedIn) as well as the mailing lists seen in \autoref{tab:mailing-lists} following the approach used in similar surveys \cite{michaelWhatNLP2023}. The mailing lists were selected to collect diverse responses from researchers in multiple fields in which the term ``intelligence'' is used, including NLP, ML, Cognitive Science, Philosophy, and Psychology. We additionally distributed the survey through internal mailing lists in research groups within the authors' networks (mostly NLP,  Computational Social Science and Human-Computer Interaction).

\section{Question Design}

This section provides a brief overview of the questions included in the survey and the rationale for including them. An overview of the questions is provided in \autoref{app:sec:questions}.

\subsection{Demographics}
\label{sec:method-demographics}
The survey includes the following background questions: the respondents' primary/secondary area of research (\textit{Q1-1.2}), career stage (\textit{Q2}), work sector (\textit{Q3}), regions of origin and workplace (\textit{Q4-5}), and gender (\textit{Q6}) following the standards of the European Social Survey \cite{europeansocialsurveyeuropeanresearchinfrastructureessericEuropeanSocialSurveyESS1120232024}. When asking for region (of origin/workplace), we used the UN geo-scheme, containing 22 sub-regions. The responses to these questions are used to describe the demographics of the survey respondents.

% \begin{table}[!t]
% \centering
% \small
% \resizebox{\columnwidth}{!}{\begin{tabular}{lr}
% \hline
% \textbf{Criterion} & \textbf{Sources} \\
% \hline
% Prob. solving & \citet{turingICOMPUTINGMACHINERYINTELLIGENCE1950, mitchellWhyAIHarderWeThink2021}  \\
% Knowledge acq. & \citet{Piaget1947PsychologieIntelligence,Chollet2019MeasureIntelligence} \\
% Adaptability & \citet{leggCollectionDefinitionsIntelligence2007, Pfeifer2006BodyIntelligence} \\
% Generalization & \citet{Marcus2003AlgebraicMind, Tenenbaum2011Mind}  \\
% Goal achievement & \citet{Piaget1952OriginsIntelligence, VanGerven2017ComputationalIntelligence} \\
% Reasoning & \citet{leggCollectionDefinitionsIntelligence2007, LakeTenenbaum2017LearnThinkPeople} \\
% Understanding & \citet{Gardner1983Mind, Tenenbaum2011Mind} \\
% Perception & \citet{Smith2005development, VanGerven2017ComputationalIntelligence} \\
% Planning & \citet{albusOutlineTheoryIntelligence1991} \\
% Creativity & \citet{Wiggins2020CreativityInformationConsciousness} \\
% Env. interaction & \citet{Piaget1952OriginsIntelligence, Pfeifer2001UnderstandingIntelligence} \\
% Consciousness & \citet{Wiggins2020CreativityInformationConsciousness} \\
% Embodiment & \citet{Smith2005development, Pfeifer2006BodyIntelligence} \\
% \hline
% \end{tabular}}
% \caption{The criteria of ``intelligence'' presented to the survey participants, and their sources (methodology described in \autoref{sec:method-criteria}).}
% \label{tab:criteria-definition}
% \end{table}

\subsection{Intelligence Criteria}
\label{sec:method-criteria}

The $> 70$ definitions of intelligence discussed in \autoref{sec:related} are unfeasible to present in a survey. Hence, we analyzed the list and additional supporting literature highlighted in \autoref{tab:criteria-definition} to identify recurring criteria. The result was a list of $10$ criteria, which we presented to the survey participants as options to \textit{Q7} (\textit{Which of the criteria are relevant for your use of the term `intelligence'?}).\footnote{The survey originally comprised $13$ criteria (see \autoref{app:all-criteria}). We limit the analysis to the $10$ well-defined terms.} The set of criteria is shown in \autoref{fig:criteria-network-graph} and \autoref{tab:criteria-definition}. The latter also lists the key sources for each criterion.

In \textit{Q9}, the respondents were asked to identify the subset of the criteria that they believe to be lacking from current LLM-based systems. In the case of both \textit{Q7} and \textit{Q9} we opted to present this as a binary choice in order to reduce the complexity of the survey. We discuss this issue in \autoref{sec:limitations}.

% As discussed in \autoref{sec:related}, there are over 70 definitions of ``intelligence'' in various fields, many of them overlapping. This would be unfeasible to present in a survey. Hence, we analyzed this list of definitions to identify recurring criteria. We also considered other relevant criteria, based on our own literature review (with sources not covered by the list of \citet{leggCollectionDefinitionsIntelligence2007}, from the fields of NLP, ML, AI, Neuroscience, Psychology, and Philosophy). The result was a list of $13$ criteria, which we presented to the survey participants as options to \textit{Q7} (\textit{Which of the criteria are relevant for your use of the term `intelligence'?}). Our full set of criteria is shown in \autoref{fig:criteria-network-graph} and \autoref{tab:criteria-definition}. The latter also lists the key sources for each criterion.

\subsection{Intelligence of LLMs}
\label{sec:method-llm-intelligence}

Consistently with \textit{Q7} and \textit{Q9}, \textit{Q8} (\textit{Do you agree that current LLM-based systems are intelligent?}) entails a binary conceptualization of ``intelligence'', with 4 Likert scale answer options (`strongly agree', `agree', `disagree', `strongly disagree'). There is a strong argument that intelligence rather exists on a spectrum \cite{dennettBacteriaBachBackEvolutionMinds2018}. However, the ``degrees of intelligence'' framing is also seen a problematic by some, as it is possible to have different kinds of cognition that do not follow a hierarchy \cite{de2016we}. We thus present \textit{Q8} as a question specifically targeting the \textit{current} LLM-based systems.
% An argument can be made that intelligence rather exists on a spectrum \cite{dennettBacteriaBachBackEvolutionMinds2018}. However, the ``degrees of intelligence'' framing is also problematic in terms of the survey as it presupposes that some entities are more intelligent than others, while it is also possible to have different kinds of cognition not following a hierarchy \cite{de2016we}. Our solution was to present \textit{Q8} as a question specifically targeting the \textit{current} LLM-based systems. We further supplemented it with \textit{Q11}: \textit{Which of the following [entities] would you consider intelligent?} The answer options for this question presented a range of biological and artificial entities (see \autoref{tab:survey-questions}).
% \footnote{Biological entities: amoebas, ants, cats, average human adults. Artificial entities: current LLM-based chatbot systems (e.g. ChatGPT), ants, current autonomous LLM-based agents (e.g. based on ChatGPT), current autonomous robotic systems (e.g. self-driving cars), current ‘narrow’ systems performing a specific task (e.g. chess, protein structure prediction), earlier chatbot system (e.g. customer support bots). The options were randomized, with `none of the above' as the final option.}

We further supplemented it with \textit{Q11}: \textit{Which of the following [entities] would you consider intelligent?} The answer options for this question presented a range of biological and artificial entities (see \autoref{tab:survey-questions}). Using a similar 4-point Likert scale, we also ask the respondents whether they believe that future systems based on a similar technology would satisfy their notion of intelligence (\textit{Q10}).

\subsection{Perspectives on the Field}
\label{sec:method-additional-qs}
Inspired by \citet{michaelWhatNLP2023}, we pose \textit{Q12}: ``\textit{To what extent is your notion of intelligence shared by other researchers in your field?}'' This allows us to estimate to what degree real disagreement on ``intelligence'' matches the perceived disagreement. We further ask the participants which of the currently used methods for evaluating LLM-based systems they consider applicable to their notion of ``intelligence'' (\textit{Q13}), and what they consider their research goals (\textit{Q14}).

\subsection{Free-form questions}
The survey contains two optional free-form text boxes, giving respondents the opportunity to describe any criteria they find missing among the options given, and to add any additional comments they may have had.

\section{Results}
\label{sec:results}

\subsection{Survey Completion}
\label{sec:completion}

%In the initial iteration of the survey, most questions could not be skipped. 
We received a total of 303 completed responses, and 86 partially complete responses. The counts of responses to each question can be found in \autoref{app:sec:questions}. 
Many participants dropped out after the section pertaining to demographic information. We believe this is an interesting finding by itself: \textbf{even researchers sufficiently interested in ``intelligence'' to volunteer for a survey may find it hard to specify what they mean by this term} (even as a multi-choice question with a relatively limited set of commonly mentioned criteria rather than full definitions).

\subsection{Demographics}
\label{sec:demographics}

\begin{figure}[!t]
    \centering
    \includegraphics[width=1\linewidth]{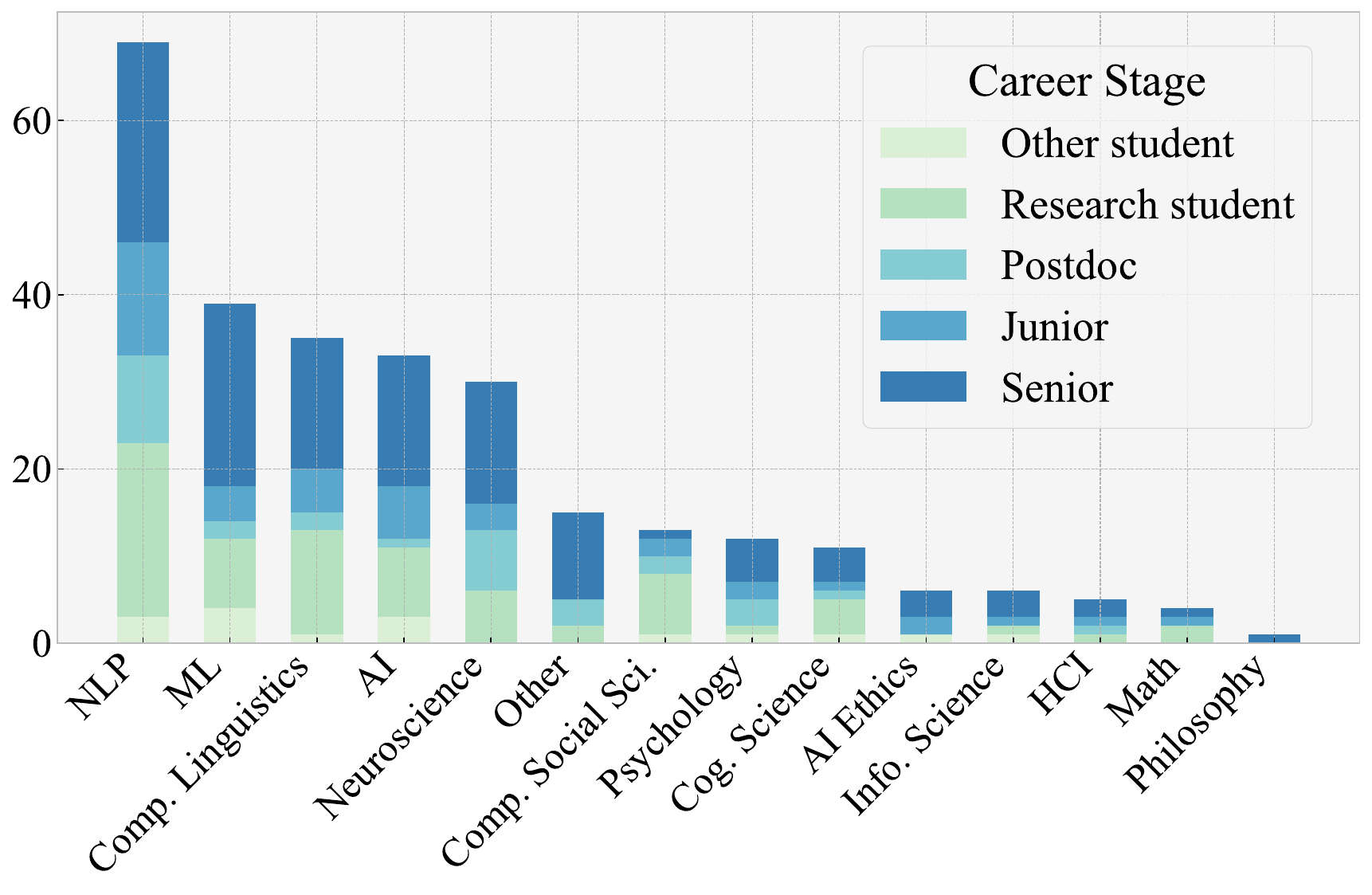}
    \caption{The number of respondents by research area.}
    \label{fig:research-area-by-stage}
\end{figure}

As shown in \autoref{fig:research-area-by-stage}, the survey respondents span a wide variety of research areas, although we received the most responses from NLP, ML, Neuroscience, and Computational Linguistics. The \textit{Other} category includes researchers from fields such as Political Science, Robotics, Economics, and Engineering.

Respondents work in a variety of geographic locations, but the majority are employed in North-Western Europe and North America. Other areas were selected more often as regions of origin than current employment (see Appendix \ref{app:fig:geo}). The occupational background of respondents was dominated by 240 respondents working in academia (including students), with an additional 59 respondents from industry, 24 working in government or non-profit organizations and an additional 8 respondents who did not wish to disclose their occupation.\footnote{The question is multi-select, as it is possible to be employed in both academia and industry simultaneously.} The survey is thus skewed towards the academic perspective. Approximately $66\%$ of respondents identified as male, $25\%$ identified as female, and the remaining $9\%$ either chose not to disclose or selected \textit{Other}.

\subsection{Intelligence Criteria}
\label{sec:sub-intelligence-criteria}

\begin{quote}
\ref{rq:def}: \textit{What are key criteria for a system to be considered ``intelligent'', and to what degree do researchers in different fields, occupations, and career stages agree?}
\end{quote}
First we assess whether there are significant group differences in the intelligence criteria selected by the respondents (see the list of criteria in \autoref{tab:criteria-definition}). We compute the \textit{phi} ($\phi$) coefficients using 2x2 contingency tables for each pair of criteria. For binary data the $\phi$ coefficient is equivalent to the Pearson correlation coefficient. This analysis thus assesses whether respondents are likely to select a criterion A if they have selected a criterion B.

\autoref{fig:criteria-network-graph} illustrates the criteria selected by respondents (nodes) as well as correlations between criteria (edges). Considering the number of respondents who selected specific criteria of intelligence, \textbf{the three top criteria  are \textit{generalization} ($\approx 86\%$), \textit{adaptability} ($\approx 83\%$), and \textit{reasoning} ($\approx 83\%$).} \textit{Reasoning} is not strongly correlated with the other two top candidates, whereas \textit{generalization} and \textit{adaptability} show the strongest correlation of all criteria ($\phi = .451$, $p \ll 0.01$).

\begin{figure}[!t]
    \centering
    \includegraphics[width=1\linewidth]{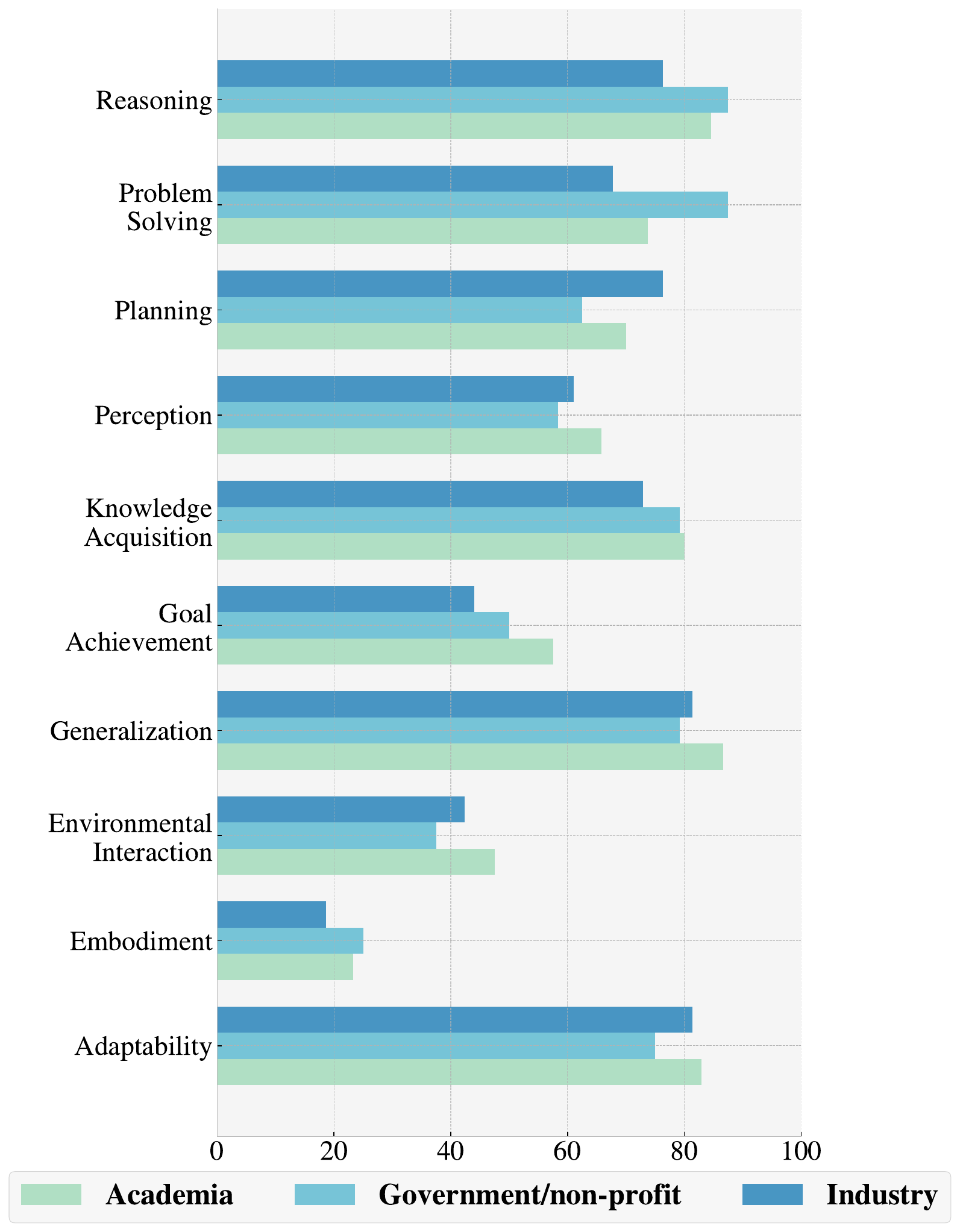}
    \caption{The percentage of respondents (x-axis) who selected a given intelligence criteria (y-axis) by primary occupation (color). \textit{Other} is removed for readability.}
    \label{fig:criteria-by-occupation}
\end{figure}

\autoref{fig:criteria-by-occupation} shows the distribution of criteria selection by occupation. To investigate the connection between selected criteria of intelligence and demographic factors we use Fisher's exact test, to account for the small sample size observed in respondent groups such as HCI, philosophy, and mathematics. We construct a contingency table with selection rates for each group and compare criterion selection rates. The strength of association is quantified via Cramer's V ($\phi_c$). \textbf{We observe no significant differences in the selection of intelligence criteria when grouping by either occupation, career stage, or primary research area}. The only exception is an effect of career stage on embodiment ($\phi_c = .218$, $p = .01$), which is driven by low selection rate of these criteria by post-docs. Likewise, we observe an effect for environment interaction ($\phi_c = .225$, $p = .003$), which is driven by its low selection rate by students. These results indicate an overall coherence in perspectives on ``intelligence'' that stands in contrast to the career stage effect on the perceived intelligence of current LLM-based systems (\autoref{sec:sub-llm-intelligence}). We also assess whether there are any geographical biases in our results and find no significant differences in criteria selection based on either origin or work region.

\begin{table}[!t]
\centering
\footnotesize
%\resizebox{\columnwidth}{!}{
\begin{tabular}{lrr}
\hline
\textbf{Criterion} & \textbf{Important} & \textbf{Lacking} \\
\hline
\multicolumn{3}{c}{\textit{\% Important overall > \% Lacking in LLMs}} \\
Reasoning               &  $83\%$  &  $59\%$ \\
Generalization          &  $86\%$  &  $54\%$ \\
Adaptability            &  $83\%$  &  $55\%$ \\
Planning                &  $70\%$  &  $58\%$ \\
Knowledge acquisition   &  $79\%$  &  $44\%$ \\
Perception              &  $63\%$  &  $47\%$ \\
Problem solving         &  $74\%$  &  $32\%$ \\
% Creativity              &  $56\%$  &  $50\%$ \\
% Understanding           &  $57\%$  &  $36\%$ \\
Env. interaction        &  $45\%$  &  $43\%$ \\
Goal Achievement        &  $55\%$  &  $32\%$ \\
%\hline
\multicolumn{3}{c}{\textit{\% Lacking in LLMs > \% Important overall}} \\
% Consciousness           &   $35\%$  &  $63\%$ \\
Embodiment              &   $23\%$  &  $53\%$ \\
\hline
\end{tabular}
%}
\caption{Percentage of respondents who selected the listed criteria as important for their notion of ``intelligence'', and who consider them lacking in the current LLM-based systems.}
\label{tab:criteria-comparison}
\end{table}

We additionally asked which of the criteria are lacking in the current LLM-based systems. \mbox{\autoref{tab:criteria-comparison}} shows the percentage of respondents who selected various criteria as important and who consider them lacking in the current systems.
For most criteria, more respondents listed them as important for their notion of ``intelligence'' than those who listed them as lacking in the current systems, which suggests that researchers are hesitant to state definitively what current systems cannot do.

For the top $3$ criteria of intelligence selected by the majority of respondents  (\textit{generalization, adaptability \& reasoning}), all of them are considered to be lacking in the current systems by 54-59\% of respondents. On the other end of the spectrum, only 32-36\% of respondents stated that the current systems are lacking in \textit{goal achievement, problem solving, \& understanding} - see \autoref{app:all-criteria}. The latter is surprising, given that many respondents come from NLP \& Computational Linguistics communities, where numerous studies and position papers have highlighted the reliance on surface-level statistical patterns as evidence against ``understanding'' \cite[][among others]{bender-koller-2020-climbing,raychoudhuryMachineReadingFastSlowWhen2022,MitchellKrakauer_2023_debate_over_understanding_in_AIs_large_language_models,wu-etal-2024-reasoning}. This number indicates a change since 2022, when a survey found an even split when participants were asked whether text-only generative models could be said to ``understand natural language in some non-trivial sense'' \cite{michaelWhatNLP2023}.

The only criterion that is considered lacking in LLMs by a higher percentage of respondents than those who consider it overall relevant for ``intelligence'' is \textit{embodiment}; a criterion that surely does not apply to systems like ChatGPT, yet, surprisingly, 47\% do not select it as lacking.

35 of the 303 respondents described additional criteria. They highlighted, in particular, social and emotional traits, and the fact that intelligence should be understood as a spectrum rather than a binary. Self-awareness and being alive or conscious was also a recurrent comment, and a few respondents emphasized abilities in knowledge-acquisition and more abstract generalization. Respondents additionally emphasized the difficulty in defining some of the criteria we use to pinpoint their notion of intelligence. A sample of free-text comments is provided in \autoref{app:sec:comments}.

\subsection{Intelligence of LLMs}
\label{sec:sub-llm-intelligence}

\begin{quote}
\ref{rq:chatgpt}: \textit{What is the perception of the ``intelligence'' of current LLM-based systems such as ChatGPT?}
\end{quote}

\begin{figure}[!t]
    \centering
    \includegraphics[width=1\linewidth]{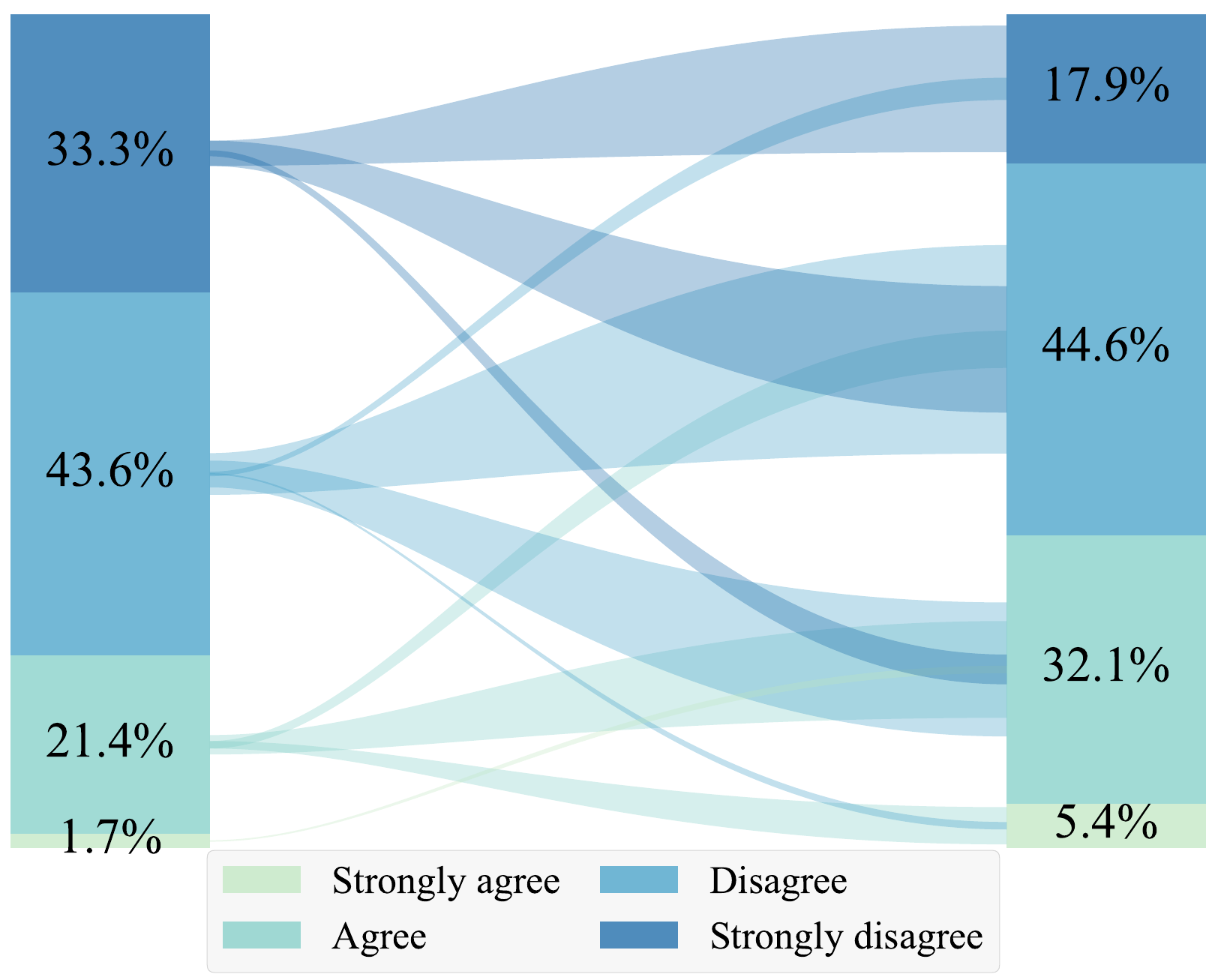}
    \caption{Sankey diagram illustrating the flow of beliefs concerning intelligence in current and future LLM based systems for senior researchers. See \autoref{app:sec:sankey} for additional groups.}
    \label{fig:intelligence-belief-development}
\end{figure}

\autoref{tab:llm-intelligence} shows that \textbf{the majority of the survey respondents do not consider the current systems intelligent} ($\approx 71\%$). When they are asked about future systems (based on a similar technology), skepticism remains the majority position ($\approx 60\%$), but there are respondents moving both from 'agree' to 'strongly agree', and from 'strongly disagree' to 'disagree'. 
The increase in strong agreement indicates a stronger belief that future LLMs will be intelligent, but the decrease in strong disagreement could also signal the unwillingness to strongly commit to a prediction of the future developments. We visualize these trends for specific groups as Sankey diagrams. \autoref{fig:intelligence-belief-development} presents these results for the senior researchers, and equivalent diagrams for other career stages are available in \autoref{app:sec:sankey}.

Next, we examine the possible effect of demographic factors on the beliefs of intelligence in the current LLM-based systems.

\begin{table}[!t]
\centering
\footnotesize
\begin{tabular}{p{1.5cm}p{1.5cm}p{1.5cm}p{1.5cm}}
\hline
\textbf{Strongly agree} & \textbf{Agree} & \textbf{Disagree} & \textbf{Strongly disagree} \\
\hline
\multicolumn{4}{c}{\textit{``Current LLM-based systems are intelligent''}} \\
$3\%$ & $27\%$ & $40\%$ & $\textbf{31\%}$ \\
\hline
\multicolumn{4}{c}{\textit{``Future systems* will be intelligent''}} \\
$7\%$ & $33\%$ & $44\%$ & $\textbf{16\%}$ \\
\hline
\end{tabular}
\caption{Difference in responses on whether current LLM-based systems or future systems (*based on a similar technology) are intelligent.}
\label{tab:llm-intelligence}
\end{table}

\paragraph{Career Stage Effect.} We observe a significant effect of career stage ($\phi_c = 0.164$, $p = 0.033$). Early-career researchers (postdocs and students) are more likely to consider the current systems intelligent ($\approx 48\%$ of postdocs agree or strongly agree, for students that number is $\approx 56\%$). Senior researchers demonstrate greater skepticism, with $\approx 77\%$ disagreeing or strongly disagreeing.
\paragraph{Research Area Effects.} We observe a significant effect of research area ($\phi_c = 0.270$, $p = 0.001$). Researchers are generally skeptical of LLM intelligence, with the majority of respondents either disagreeing or strongly disagreeing in every research area except for HCI ($60\%$ agree that LLM-based systems are intelligent). However, we have relatively few responses from HCI researchers (8). Following HCI, researchers from Cognitive Science and CSS are the most positive regarding LLM intelligence, both with $\approx 45\%$ positive responses. Interestingly, the most skeptical groups are the respondents who selected ``AI'' or ``Computational Linguistics''. The vast majority in both groups did not agree that ``current systems are intelligent'' ($\approx 73\%$ and $\approx86\%$ respectively).

We conducted a similar analysis with respect to views on the intelligence of future systems, and found no significant differences among demographic groups.

\begin{figure}[!t]
    \centering
    \includegraphics[width=1\linewidth]{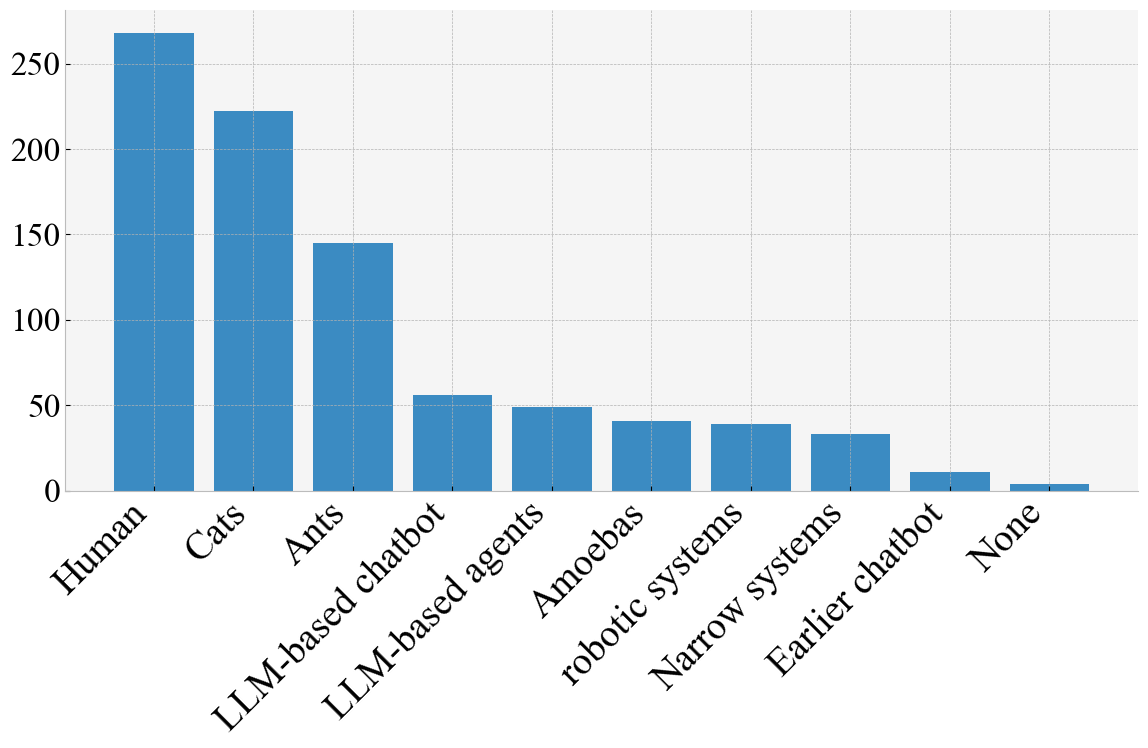}
    \caption{Number of respondent selecting an entity when asked whether the entity should be considered intelligent.}
    \label{fig:intelligent-entities}
\end{figure}

\paragraph{Are LLMs as intelligent as other entities?} 
\label{sec:sub-intelligent-entities}
As discussed in \autoref{sec:method-llm-intelligence}, we opted for a binary perspective on whether the current LLM-based systems are intelligent. However, we also provided the respondents with the opportunity to rate a wide range of biological and artificial systems as intelligent or not. The distribution of their answers is presented in \autoref{fig:intelligent-entities}. Almost all respondents stated that humans are intelligent. All biological entities except amoebas have more respondents rating them as intelligent, than all artificial systems. Within the latter, the LLM-based systems are selected as intelligent more often than others (though the difference is not as significant as between the biological entities). In the order of number of researchers rating various entities as intelligent, \textbf{the current LLM-based systems rank between ants and amoebas.} Interestingly, the agentic systems have slightly fewer supporters than the non-agentic ones. 

\subsection{Perspectives on the Field}
\label{sec:perspectives-results}

\begin{quote}
\ref{rq:agenda}: \textit{What role, if any, does ones research agenda play in the notion of ``intelligence''?}
\end{quote}

\paragraph{Research Agenda.}
Following the same methodology as used in \autoref{sec:sub-intelligence-criteria}, we assess whether the research agenda of the respondents influences their selection of intelligence criteria and their perception of intelligence of LLM-based systems. We observe no significant differences with regards to the former. For the latter, we find that the researchers focusing on the interaction of tech and society are more likely to `strongly disagree' with the statement that these systems are intelligent (for the current systems $\phi_c = .248$, $p = .001$, for the future systems $\phi_c =  .187$, $p = .032$). Likewise, \textbf{the  researchers who aim at creating intelligent systems are more likely to consider LLM-based systems intelligent} (for the current systems $\phi_c = .208$, $p = .004$, for future systems $\phi_c = .216$, $p = .01$). This is an interesting observation, given the above finding that the respondents identifying their primary field of research as ``AI'' are among the most skeptical about the current systems. This suggests a dissociation between ``AI'' as a research field and ``intelligent technology'' as a research goal.

\begin{table}[!t]
\centering
\small
\resizebox{\columnwidth}{!}{\begin{tabular}{lr}
\hline
\textbf{Research Goal} & \textbf{Selection rate} \\
\hline
Adding to Scientific Knowledge & $72.5\%$ \\
Creating Intelligent Technology & $16.2\%$ \\
Creating Practically Useful Technology &  $58.1\%$ \\
Discussing Tech / Societal Interaction &  $21.8\%$ \\
Other &  $4.6\%$ \\
\hline
\end{tabular}}
\caption{Overview of respondent research goals. Respondents were allowed to select multiple options.}
\label{tab:research-goals}
\end{table}

\paragraph{Perception of others' beliefs.}

We find that \textbf{\mbox{$\approx 45\%$} of respondents believe that more (or most) of their peers would agree than disagree with their notion of ``intelligence''}. A further $\approx 35\%$ are unsure about the perspective of their peers. A minority of the respondents believe that their notion of ``intelligence'' is in the minority ($\approx 20\%$).

\paragraph{Tests of Intelligence.}

\textbf{The vast majority of respondents (62\%) do not believe that we have a test adequately measuring ``intelligence''.} The respondents generally rejected all listed options (including composite benchmarks, Turing test, IQ and professional knowledge tests). The option that had the most support was generalization tests ($37\%$).

\paragraph{Feedback on the survey and additional comments.}

We received 33 unstructured comments. A part of them was feedback on the survey. 
Some respondents commented that the question options were too forced, that intelligence is a spectrum (see \autoref{sec:method-llm-intelligence}), and a few argued that intelligence is not a useful concept for assessing neither humans nor systems. We also received more general comments towards the discussion of intelligence (see \autoref{app:sec:comments} for a sample of free-text comments).

\section{Discussion}
\label{sec:discussion}

This survey provides concrete numbers to a core source of confusion in the field: what we mean when we use the term ``intelligence''. While most papers do not define the term, and many researchers struggle to specify what they mean (\autoref{sec:completion}), we do observe a degree of coherence. The majority of respondents ($> 80 \%$), regardless of their career stage, primary research area and occupation, agree that ``intelligence'' relies on the criteria of \textit{generalization}, \textit{adaptability} and \textit{reasoning} (\ref{rq:def}). The high agreement on these criteria offers the community an opportunity to make the discussion of ``intelligence'' more specific, and encourage evaluation that focuses on them. This is also consistent with the fact that generalization tests, such as the ARC challenge \cite{Chollet2019MeasureIntelligence}, was the most selected among respondents ($\approx 37\%$). % and that the respondents were skeptical of current evaluation techniques similarly to the findings of the AAAI's survey on the future of AI \cite[][p. 25]{rossiFutureAIResearch2025}

Given that there is some consistency to the core criteria of ``intelligence'', we would expect significant agreement on whether this term applies to the current LLM-based systems (\ref{rq:chatgpt}). Indeed, this is what we find: the majority of respondents are skeptical about both current ($\approx 71\%$)  and future systems based on a similar technology ($\approx 60\%$) (\autoref{sec:sub-llm-intelligence}).\footnote{A report from `The Association for the Advancement of AI' on the future of AI found a similar skepticism regarding neural approaches being enough to achieve general-purpose intelligence matching that of humans \cite[][p. 74]{rossiFutureAIResearch2025}.} This skepticism stems in part from the limitations of current systems to generalize and reason (\autoref{tab:criteria-comparison}). Given that most respondents consider these criteria core to the notion of ``intelligence'', they would be expected to be skeptical of definitions that do not rely on these criteria, such as OpenAI reportedly tying a definition of `AGI' to profit yields \cite{Zeff_2024_Microsoft_and_OpenAI_have_financial_definition_of_AGI_Report}.

% \footnote{Notably, OpenAI was reported to define `AGI' in terms of profit that this technology can yield \cite{Zeff_2024_Microsoft_and_OpenAI_have_financial_definition_of_AGI_Report}.}
%We also find that, given a range of various biological and artificial entities that could be considered ``intelligent'', the researchers rate the LLM-based systems as ``intelligent'' more often than amoebas, but less often than ants (\autoref{fig:intelligent-entities}).

Finally, the results of this survey provide context for the recent debate on the relation between the fields of ``NLP'' and ``AI'' \cite{Bender_2024_ACL_Is_Not_AI_Conference,Goldberg_2024_ACL_is_not_AI_Conference_,Mortensen_2024_Is_ACL_AI_or_NLP_or_CL_Conference}. The majority of survey respondents do not see their goal as creating intelligent systems, focusing instead on advancing scientific knowledge. However, we found an interesting correlation between research goals and beliefs regarding the intelligence of current systems (\ref{rq:agenda}). The respondents who do aim to create intelligent systems were more likely to attribute intelligence to both the current and future LLM-based systems. %&and they do not necessarily identify with ``AI'' as a research field ($\approx 29\%$ in AI, $\approx 25\%$ in NLP, and $\approx 17\%$ in ML).

\section{Conclusion}
\label{sec:conclusion}

The term ``intelligence'' is often used, but rarely defined in the current research on LLMs. Despite the lack of clear definitions, we find a high degree of consensus across research fields that \textit{generalization}, \textit{adaptability}, and \textit{reasoning} are key to the notion of ``intelligence''. The majority of the survey respondents are skeptical of applying this term to the current and future systems based on LLMs. We find that the senior researchers tend to be more skeptical, and respondents who view creation of  intelligent systems as part of their research agenda are more likely to attribute intelligence to both current and future systems. 

To the best of our knowledge, this is the most comprehensive effort to date to clarify what the experts currently mean by the term `intelligence', and whether it applies to the current technology. These results provide a useful point of reference for both public discourse and evaluation of marketing claims.

%Our results highlight a discrepancy between how researchers conceptualize intelligence and how this term is used in public discourse and marketing.
\section*{Limitations}
\label{sec:limitations}
\paragraph{Comprehensiveness.} Defining ``intelligence'' is in many ways an impossible task even in a long-form text. This study covered only a small subset of criteria that can be found in existing definitions of ``intelligence'' (and for which the scope of this study would not allow for a full literature review). As discussed in \autoref{sec:method-criteria}, we aimed to present at least the core criteria, and most of the criteria we chose are commonly mentioned in the current literature on LLM-based systems.

\paragraph{Intelligence as a binary attribute.} To keep the survey question complexity manageable, we had to simplify many questions, including the binary choice of different criteria as either present or not present in different systems. Given the difficulty that our respondents had even with this design (\autoref{sec:results}), we recommend that future work on this issue would consider either more focused surveys that ask fewer questions in greater detail, or a different methodology. 

Our Likert scale questions about applicability of the term ``intelligence'' to the current/future LLM-based systems could present it as a variable on a spectrum, but we opted for the binary presentation for the reasons outlined in \autoref{sec:method-llm-intelligence}. 

\paragraph{Assumption of coherent notion of ``intelligence''.} Our survey targets the use of the term ``intelligence'' by researchers, on the assumption that for each individual there is a coherent notion underlying their use of this term. It is possible that this assumption is false, and researchers use this term differently for different entities (cf. the proposal to consider LLMs as ``a new type of intelligence'' by \citet{doi:10.1073/pnas.2322420121}). This possibility necessitates further research.

\paragraph{Definitions of intelligence criteria.} Many of the criteria we listed, especially ``reasoning'' and ``understanding'', are themselves lacking clear definitions, and merit similar surveys. Hence, while we can conclude that respondent views on intelligence criteria are highly coherent, we cannot be certain that each of the $13$ intelligence criteria have the same definition for every respondent.

\paragraph{Representativeness.} The survey respondents necessarily present only a sample of the researcher population, which poses questions of its representativeness. As discussed in \autoref{sec:demographics}, our sample is skewed towards academics from the ``western'' world, and may not represent the views of researchers in other regions and organization types. To mitigate that, we recommend that future work should identify and reach out to distribution channels beyond the mailing lists and social media based in the West.

\paragraph{Selection bias.} We acknowledge the possibility that since the respondents sample is affected by self-selection bias, drawing the attention of the researchers who are more interested in ``intelligence'' than average. Given that, our result of 16.2\% respondents who view intelligent systems as a research goal may be an overestimate.
\section*{Ethical Considerations}
\label{sec:ethics}

\paragraph{Broader impact}
This study maps the criteria that are believed by researchers to be important for building intelligent systems. The survey merely provides a snapshot of the sentiment in `AI' related research communities, but we hope that our results will help frame the discussion on what intelligence is and how to develop intelligent systems. It is further the hope that our survey can serve as a useful point of comparison for future studies in a field that is evolving at a rapid pace.

\paragraph{Personal and sensitive information}
All responses were completely anonymous and thus designed to not solicit any personally identifiable information or fine-grained demographic information about participants. Since no personal or contact information was collected, it was not possible to provide the respondents with the option to withdraw or alter their responses after the completion of the survey.

Due to low response-rates for certain subgroups, the publicly available dataset is curated to remove some demographic information. In particular, we remove the gender data, and merge the subdivisions of geographical regions, so that e.g. both Eastern and Middle Africa are coded as ``Africa''.).

\paragraph{Data and code availability}
All code used to perform statistical tests and report results, as well as survey data and full questionnaire is included in the supplementary materials, and will be publicly available under CC BY-NC 4.0 license upon the publication of this study.

\paragraph{Competing interests} The authors have no relevant financial or non-financial interests to disclose.

\paragraph{Ethics Statement} The study was approved by the Research Ethics Committee at the authors' institution.
%\section*{Acknowledgments}

%Acknowledgments will appear in camera-ready version.

% Rob van der Goot, Elisa Bassignana, NLPnorth group

\bibliography{references,manual_refs,anthology}

\appendix
\clearpage

\section{Appendix: Additional Criteria}
\label{app:all-criteria}

This section provides a brief analysis of the analysis with the full set of criteria (those seen in \autoref{tab:criteria-definition} + \autoref{tab:criteria-definition-additional}). We find significant differences in the selection of \textit{consciousness} based on career stage ($\phi_c = .215$, $p = .006$). The results rely on low sample size and we thus do not take them to be reliable.

\begin{table}[!h]
\centering
\small
\resizebox{\columnwidth}{!}{\begin{tabular}{lr}
\hline
\textbf{Criterion} & \textbf{Sources} \\
\hline
Understanding & \citet{Tenenbaum2011Mind} \\
Creativity & \citet{Wiggins2020CreativityInformationConsciousness} \\
Consciousness & \citet{Wiggins2020CreativityInformationConsciousness} \\
\hline
\end{tabular}}
\caption{The additional criteria of ``intelligence'' presented to the survey participants, and their sources (methodology described in \autoref{sec:method-criteria}). These three remaining criteria were removed from the main analysis due to a lack of clear definitions of the terms.}
\label{tab:criteria-definition-additional}
\end{table}

\begin{table}[!h]
\centering
\footnotesize
%\resizebox{\columnwidth}{!}{
\begin{tabular}{lrr}
\hline
\textbf{Criterion} & \textbf{Important} & \textbf{Lacking} \\
\hline
\multicolumn{3}{c}{\textit{\% Important overall > \% Lacking in LLMs}} \\
Creativity              &  $56\%$  &  $50\%$ \\
Understanding           &  $57\%$  &  $36\%$ \\
\hline
\multicolumn{3}{c}{\textit{\% Lacking in LLMs > \% Important overall}} \\
Consciousness           &   $35\%$  &  $63\%$ \\
\hline
\end{tabular}
%}
\caption{Percentage of respondents who selected the listed criteria as important for their notion of ``intelligence'', and who consider them lacking in the current LLM-based systems.}
\label{tab:criteria-comparison-additional}
\end{table}

\begin{figure}[!h]
    \centering
    \includegraphics[width=.9\linewidth]{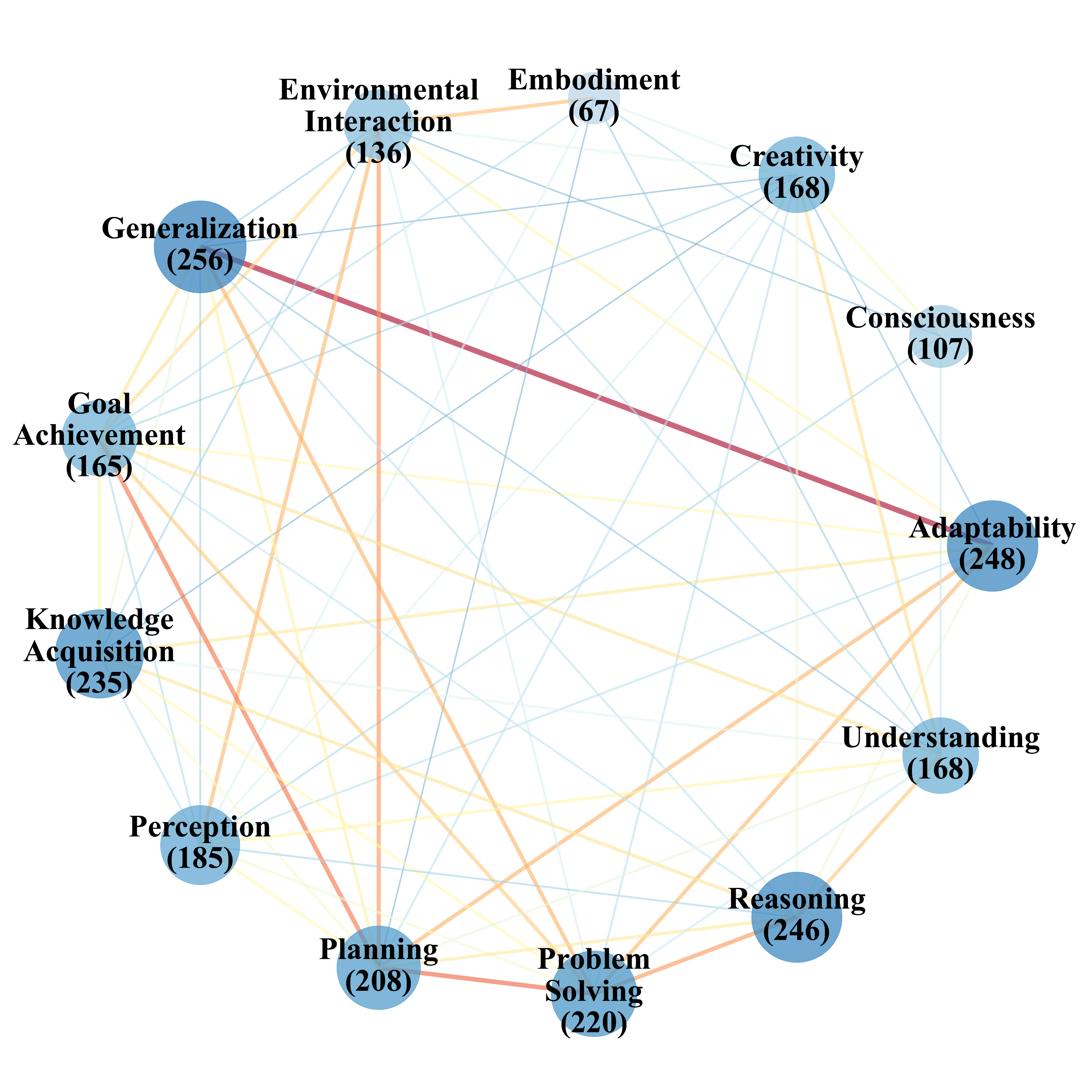}
    \caption{Correlation between criteria that the survey respondents selected as relevant for their notion of ``intelligence''. Darker edges indicate stronger correlations, larger nodes indicate higher relevance. Only edges with $\phi > |0.1|$ are shown.}
    \label{fig:criteria-network-graph-all}
\end{figure}

\newpage

\section{Appendix: Survey Questions}
\label{app:sec:questions}

\autoref{tab:survey-questions} provides a full list of survey questions together with answer options. The survey is also available as a form in the supplementary materials.

\begin{table*}[!h]
\footnotesize
\centering
\begin{adjustbox}{max width=\textwidth}
\begin{tabularx}{\textwidth}{lXp{1.3cm}}
\hline
\textbf{ID} & \textbf{Question} & \textbf{no. of responses} \\
&& \\
\hline
\rowcolor{lightgray} & \centering\textbf{Demographics related} & \\
\hline
\textbf{Q1} & What area of research do you consider your primary area? & 303 \\
\textbf{Q1.2} & If you also work in another area please select that: & 256 \\
& Answer options for Q1-1.2: \textit{AI Ethics/Governance, Artificial Intelligence, Computational Linguistics, Computational Social Science, Cognitive Science, Human-Computer Interaction, Information Science, Machine Learning, Mathematics, Natural Language Processing, Neuroscience, Philosophy, Psycholinguistics, Psychology (cognitive, developmental...), Other} & \\
\textbf{Q2} & What is your current career stage? & 303 \\
& 
Answer options: \textit{Senior (faculty or industry), Junior (faculty or industry), Postdoc, Research student (PhD, MPhil), 
Other student (Master, Bachelor), Other, Prefer not to say} & \\
\textbf{Q3} & In what sector do you currently work? & 303 \\
& Answer options: \textit{Academia, Industry, Government/non-profit, Prefer not to say}  & \\
\textbf{Q4} & What is your region of origin? & 303 \\
\textbf{Q5} & In what region is your primary place of work? & 303 \\
& Answer options for Q4-5: 22 regions from the UN geoscheme (\url{https://en.wikipedia.org/wiki/United_Nations_geoscheme}) & \\
\textbf{Q6} & Which of the [gender] options best describe you? & 303 \\
& Answer options: \textit{A woman, A man, Other, Prefer not to say, Don't know} & \\
&& \\
\hline
\rowcolor{lightgray} & \centering\textbf{Intelligence related} & \\
\hline
\textbf{Q7} & Which of the criteria are relevant for your use of the term `intelligence'? & 302 \\
& Answer options: \textit{Interaction with Environment (physical or virtual), Embodiment (being situated in a physical environment), Perception (extracting and acting upon useful information from the environment), Knowledge acquisition (learning, understanding or gaining knowledge and skills), Problem solving (in a familiar domain), Goal Achievement (accomplishing defined objectives and optimizing for performance on specific tasks), Reasoning (logical inference - deductive, abductive etc.), Planning (anticipating future events and organizing actions based on a deliberate strategy), Adaptability (making sense of new environments and/or handling novel tasks), Generalization (successfully handling new types of data and situations), Creativity  (in your definition of this term), Consciousness  (in your definition of this term), Language understanding  (in your definition of this term)} & \\
\textbf{Q8} & Do you agree that current LLM-based system are intelligent? & 300 \\
& Answer options: Strongly disagree, disagree, agree, strongly agree & \\
\textbf{Q9} & Which of the criteria of intelligence are lacking in the current LLM based systems (such as ChatGPT)? & 298 \\
& Answer options: same as for Q7 & \\
\textbf{Q10} & If current LLM-based systems do not satisfy your criteria for intelligence, do you agree that future systems based on similar technology will? & 291 \\
& Answer options: \textit{Strongly disagree, disagree, agree, strongly agree} & \\
\textbf{Q11} & Which of the following [entities] would you consider intelligent? & 299 \\
&Answer options: \textit{Average human adults , Cats, Ants, Amoebas, Current ‘narrow’ systems performing a specific task (e.g. chess, protein structure prediction), Current autonomous robotic systems (e.g. self-driving cars), Earlier chatbot system (e.g. customer support bots), Current LLM-based chatbot systems (e.g. ChatGPT), Current autonomous LLM-based agents (e.g. based on ChatGPT), None of the above} & \\
\textbf{Q12} & To what extent, is your notion of intelligence shared by other researchers in your field? & 298 \\
& Answer options: \textit{Most would agree with me, More would agree than disagree, More would disagree than agree, Most would disagree with me, Unsure }& \\
\textbf{Q13} & Which of the following tests, if any, do you believe measure intelligence & 299 \\
& Answer options: \textit{Human preference ranking of outputs of different models, Tests for measuring human intelligence, such as standardized IQ tests , Tests for measuring human professional knowledge, e.g. SAT or medical exam questions, Average scores on composite benchmarks for LLMs such as MMLU or BIG-bench, The Turing test, Generalization tests (focusing on differences between training and test distributions), Other (please specify), I do not believe that we currently have such a test} & \\
\textbf{Q14} & Which of the following options are the best description of your research goals? & 296 \\
&Answer options: \textit{Creating technology that qualifies for my notion of `intelligence', Creating technology that is practically useful, Documenting/critiquing the interaction between technology and the society/ecosphere, Adding to the knowledge/scientific understanding of the phenomenon I study (e.g. cognition, language, technology, society etc.), Other} & \\
&& \\
\hline
\end{tabularx}
\end{adjustbox}
\caption{Overview of questions as phrased in the survey and number of respondents who completed each question (out of 303 respondents who made it to the end of the survey). The questions with multi-choice answer options were presented in the randomized order, with ``Other'' position fixed at the end.}
\label{tab:survey-questions}
\end{table*}

%\clearpage

\section{Appendix: Mailing lists}

\autoref{tab:mailing-lists} provides a full list of mailing lists to which the survey was sent.

\begin{table*}[h]
%\begin{threeparttable}
\footnotesize
    \centering
    \begin{tabularx}{\textwidth}{l|X}
        \hline
        \textbf{Mailing List} & \textbf{Link} \\
        \hline
         ML News &  \url{https://groups.google.com/g/ml-news} \\
         Corpora List & \url{https://list.elra.info/mailman3/hyperkitty/list/corpora@list.elra.info/} \\
         EUcog News & \url{https://groups.google.com/g/eucog-general-news} \\
         Sys. Neuroscience & \url{https://groups.google.com/g/systems-neuroscience} \\
         Neural Ensemble & \url{https://groups.google.com/g/neuralensemble} \\
         Connectionist & \url{https://mailman.srv.cs.cmu.edu/mailman/listinfo/connectionists} \\
         ACL portal & \url{https://www.aclweb.org/portal/} 
    \end{tabularx}
    \caption{Overview of mailing lists used to distribute the survey.}
    \label{tab:mailing-lists}
%    \end{threeparttable}
\end{table*}

%\clearpage

\section{Appendix: Geographic Distribution of Respondents}
\label{app:fig:geo}

\begin{figure*}[!h]
    \centering
    \begin{subfigure}[b]{0.48\textwidth}
        \centering
        \includegraphics[width=\textwidth]{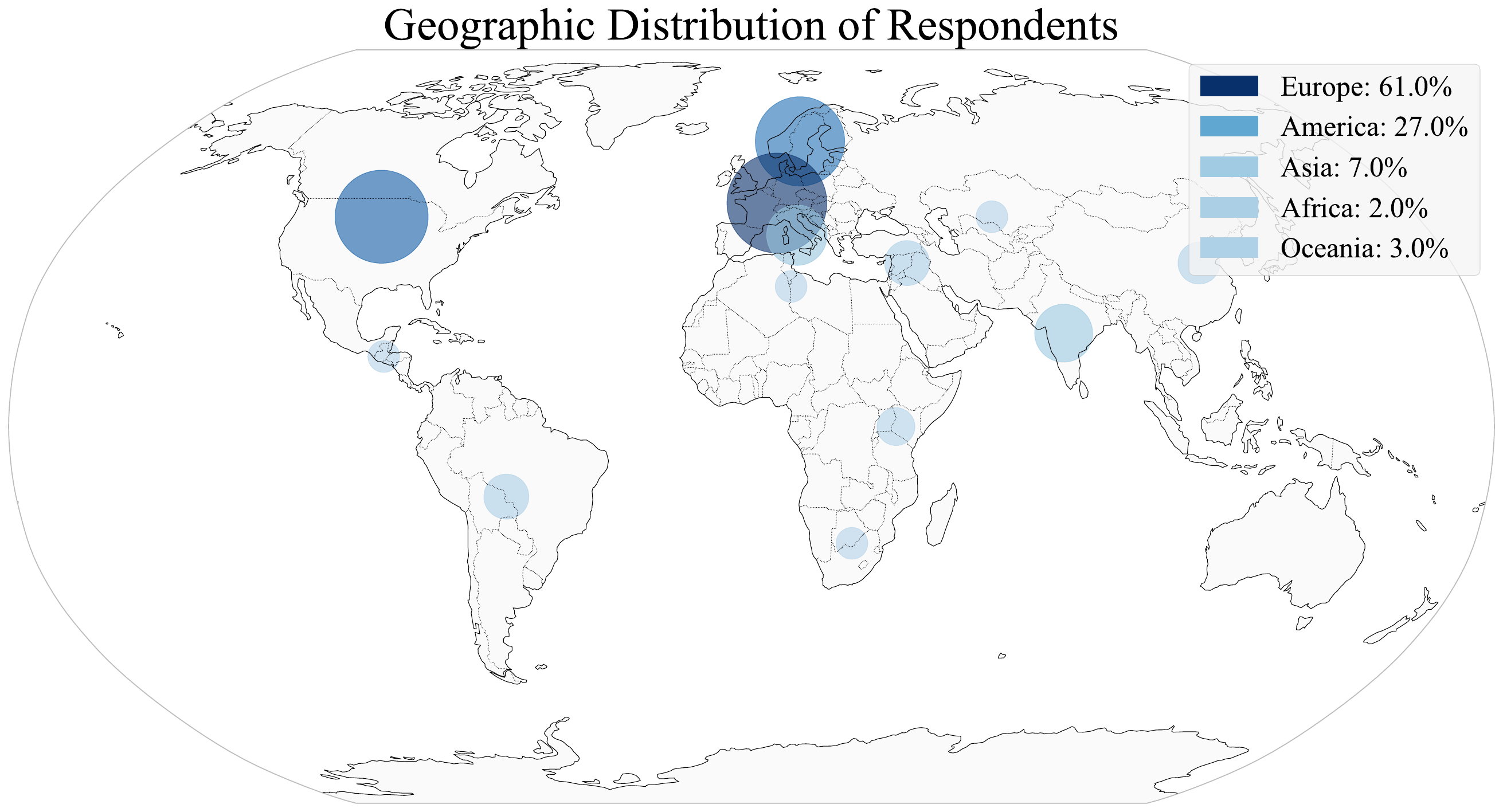}
        \caption{Geographical distribution of respondents based on their region of work.}
        \label{fig:geographical-distribution-work}
    \end{subfigure}
    \hfill
    \begin{subfigure}[b]{0.48\textwidth}
        \centering
        \includegraphics[width=\textwidth]{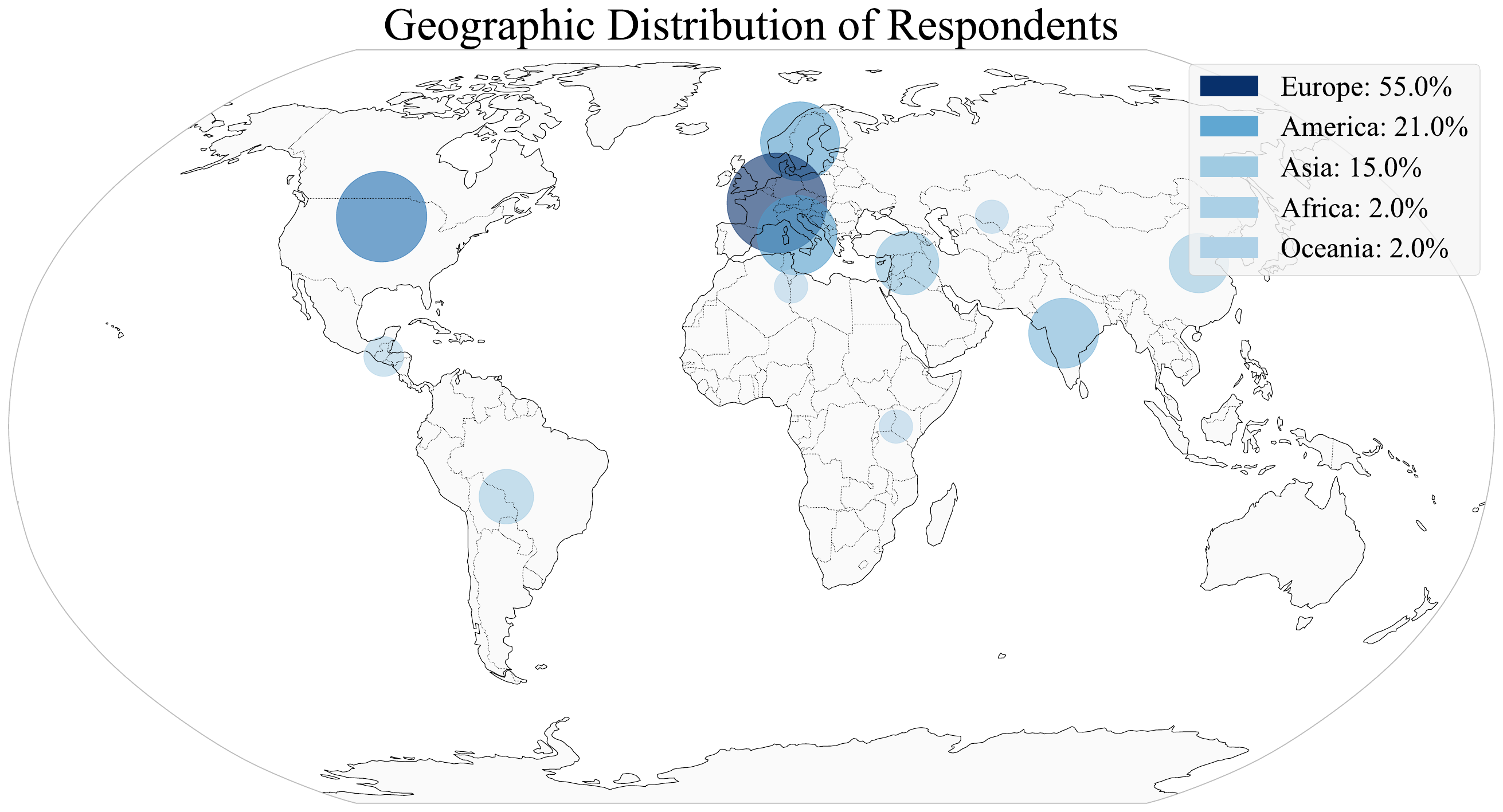}
        \caption{Geographical distribution of respondents based on their region of origin.}
        \label{fig:geographical-distribution-origin}
    \end{subfigure}
    \caption{Geographical distribution of survey respondents. Most respondents are from the ``western'' world, although more respondents are originally from e.g. Asia and Africa than currently work in those regions.}
    \label{fig:geographical-distributions}
\end{figure*}

\autoref{fig:geographical-distributions} provides an overview of the geographical distribution of respondents.

%\clearpage

\section{Appendix: Sankey Diagrams: Belief Development}
\label{app:sec:sankey}

\begin{figure*}[!h]
    \centering
    \begin{subfigure}[b]{0.48\textwidth}
        \centering
        \includegraphics[width=\textwidth]{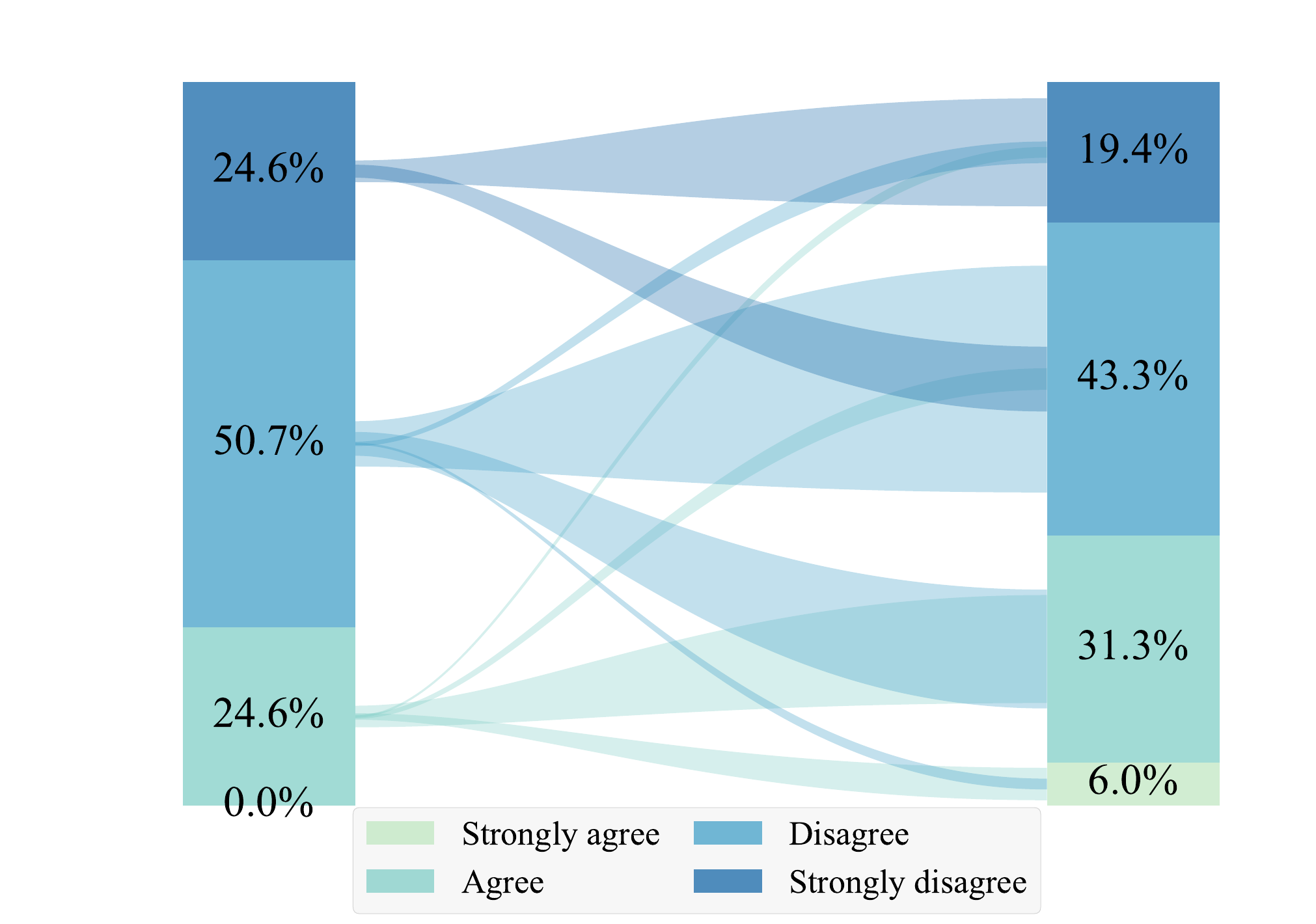}
        \caption{Sankey diagram showing changing beliefs for ``NLP'' researchers.}
        \label{fig:sankey-nlp}
    \end{subfigure}
    \hfill
    \begin{subfigure}[b]{0.48\textwidth}
        \centering
        \includegraphics[width=\textwidth]{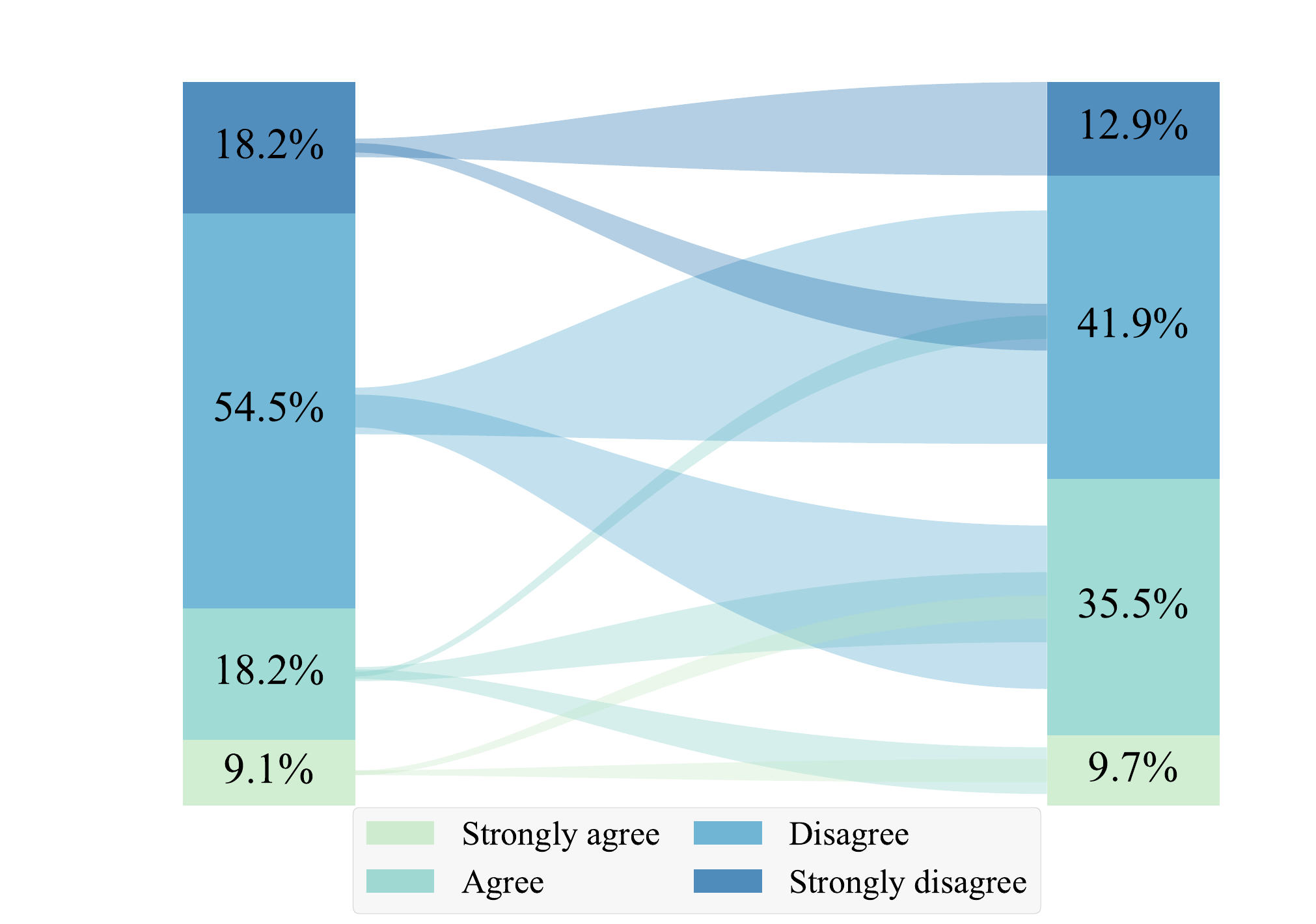}
        \caption{Sankey diagram showing changing beliefs for ``AI'' researchers.}
        \label{fig:sankey-ai}
    \end{subfigure}
    
    \vspace{1em}
    \begin{subfigure}[b]{0.48\textwidth}
        \centering
        \includegraphics[width=\textwidth]{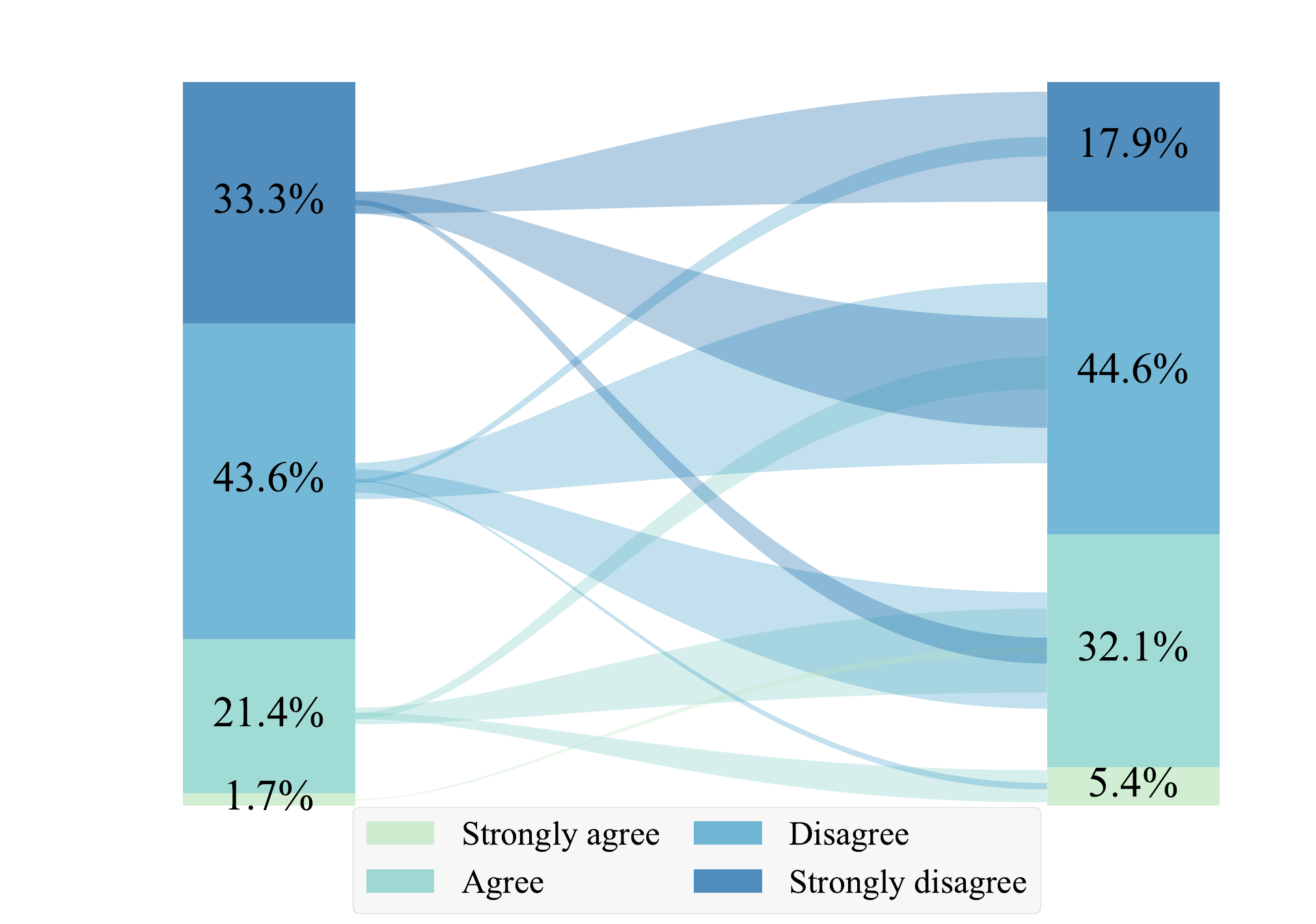}
        \caption{Sankey diagram showing changing beliefs for senior researchers.}
        \label{fig:sankey-senior}
    \end{subfigure}
    \hfill
    \begin{subfigure}[b]{0.48\textwidth}
        \centering
        \includegraphics[width=\textwidth]{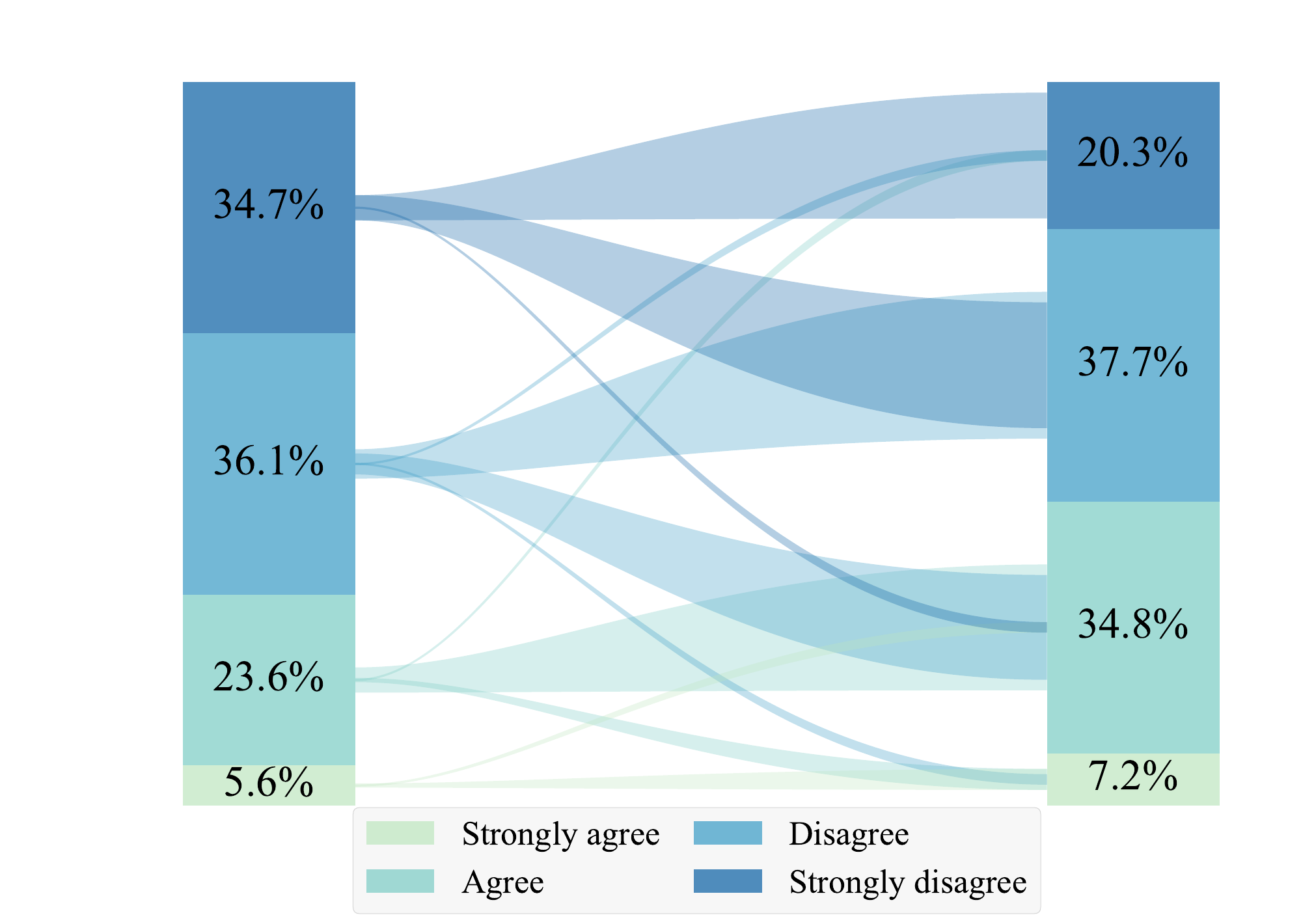}
        \caption{Sankey diagram showing changing beliefs for student researchers (PhD level).}
        \label{fig:sankey-phd}
    \end{subfigure}
    \caption{Sankey diagrams showing the different beliefs of researchers regarding the intelligence of current LLM-based systems (on the left) and future systems based on similar technology (on the right).}
    \label{fig:sankey-all}
\end{figure*}

\autoref{fig:sankey-all} showcases the beliefs of subgroups the respondents regarding the intelligence of LLM-based systems.

\begin{table*}[!h]
%\begin{threeparttable}

\footnotesize
\begin{tabularx}{\textwidth}{lX}
\hline
\textbf{Comment ID} & \textbf{Text} \\
\hline
\textbf{C1} & Emotional capacity, understanding and feeling a sense of qualia, and having some sense of humanness/mammalness that invokes/evokes responses similar to carbon based life-forms. I'm keeping this to Carbon based emulation because other loose forms of consciousness becomes very unbounded and we don't have sufficient tools or language yet to describe or understand it. \\
\hline
\textbf{C2} & Abstraction: being able to make a new concept that abstracts the core similarities of the underlying group. If using LLMs as an example, then imagine a vector that is able to be decoded into a new never before seen token.    This is to be differentiated from generalisation. \\
\hline
\textbf{C3} & Many of these are relevant to ``intelligence'', but they each sit on a spectrum. So determining ``intelligence'' isn't generally going to be a box-ticking exercise. \\
\hline
\textbf{C4} & The notion of ``intelligence'' itself is deeply problematic, with origins in eugenics and racism. The idea that people (and other beings) can be ranked according to such a property is abhorrent. The survey did not let me choose none of them, so I chose some, but under duress. \\
\hline
\textbf{C5} & I've answered this under what I think of as "weak intelligence," where consciousness isn't required but some abstract reasoning is required. \\
\hline
\end{tabularx}
\caption{Selected free-text comments from the survey participants that focus on the criteria and definitions of intelligence.}
\label{tab:criteria-comments}
%\end{threeparttable}
\end{table*}

%\clearpage

\begin{table*}[!b]
%\begin{threeparttable}
\footnotesize
\begin{tabularx}{\textwidth}{lX}
\hline
\textbf{Comment ID} & \textbf{Text} \\
\hline
\textbf{C1} & Some of the questions forced me to an oversimplification. I understand that this is, to some extent, necessary to produce immediate answers, and you cannot expect a treatise on the topic from every participant. However, adding more ``unsure'' or ``it depends'' options could have helped. \\
\hline
\textbf{C2} & Because each dimension of intelligence is a matter of degree, I can't answer yes/no questions about whether something is intelligent despite your efforts to force me to do so. Hence, most of the questions on this page are poorly formulated. LLMs, for example, are more broadly knowledgeable and have better language comprehension than any previous system. They are more intelligent along these dimensions than other systems. But there is no threshold that separates "intelligent" from "unintelligent". \\
\hline
\textbf{C3} & To me, intelligence is a spectrum -- humans and other animals can be more or less intelligent, and it is difficult to draw a line. I think this spectrum is, by definition, centered around *human* intelligence, and so it is easier to place something on the spectrum if it behaves more like humans. LLMs do not behave in a very human-like manner, so it is difficult to measure their intelligence. \\
\hline
\textbf{C4} & I actually think that ``intelligence'' is not a useful concept to assess human and animal cognition, not is it a useful concept to assess AI systems. Human and animal cognition have quite specific properties that cover a range of needs (orienting action at objects/other beings, social competence, learning from experience, etc). These can be assessed in ways known from psychology and neuroscience. Artificial agents typically have much narrower ranges of skills or properties. It may be misleading to measure and compare these to organisms through a single concept. \\
\hline
\textbf{C5} & Systems that pass the Turing test may not actually BE intelligent, but maybe the distinction between intelligence and not becomes meaningless at a certain point?  Awareness, empathy and altruism are key elements in (my interpretation of) intelligence, and I seriously doubt that we will see all three traits combined in a silicon 'mind' anytime soon. \\
\hline
\end{tabularx}
\caption{Selected free-text comments from the respondents on the general considerations on the criteria, on how to think about ``intelligence'', and general concerns regarding the discussion.}
\label{tab:general-comments}
%\end{threeparttable}
\end{table*}

\section{Appendix: Free-text Comments}
\label{app:sec:comments}

\autoref{tab:criteria-comments} and \autoref{tab:general-comments} provide a sample of comments from survey respondents.

\subsection{Selected Comments on Criteria of Intelligence}
\label{app:sec:criteria-comments}

\subsection{Selected General Comments on the Survey}
\label{app:sec:general-comments}

\end{document}